# Nearest Descent, In-Tree, and Clustering

Teng Qiu, Kai-Fu Yang, Chao-Yi Li, and Yong-Jie Li

**Abstract**—In this paper, we propose a physically inspired graph-theoretical clustering method, which first makes the data points organized into an attractive graph, called In-Tree, via a physically inspired rule, called Nearest Descent (ND). In particular, the rule of ND works to select the nearest node in the descending direction of potential as the parent node of each node, which is in essence different from the classical Gradient Descent or Steepest Descent. The constructed In-Tree proves a very good candidate for clustering due to its particular features and properties. In the In-Tree, the original clustering problem is reduced to a problem of removing a very few of undesired edges from this graph. Pleasingly, the undesired edges in In-Tree are so distinguishable that they can be easily determined in either automatic or interactive way, which is in stark contrast to the cases in the widely used Minimal Spanning Tree and k-nearest-neighbor graph. The cluster number in the proposed method can be easily determined based on some intermediate plots, and the cluster assignment for each node is easily made by quickly searching its root node in each sub-graph (also an In-Tree). The proposed method is extensively evaluated on both synthetic and real-world datasets. Overall, the proposed clustering method is a density-based one, but shows significant differences and advantages in comparison to the traditional ones. The proposed method is simple yet efficient and reliable, and is applicable to various datasets with diverse shapes, attributes and any high dimensionality.

**Index Terms**—Physically inspired, graph-theoretical, nearest descent, In-Tree, nearest ascent density clustering.

---◆---

## 1 INTRODUCTION

CLUSTERING, or cluster analysis, aims at classifying the samples in a dataset into groups based on similarities. It has been widely used by scientists and engineers to analyze multivariate data generated in diverse fields, e.g., microarray gene expression samples in biology, climate data in earth science and the massive documents or images on internet [1]. Unlike the supervised learning task such as classification which requires considerable labeled samples, clustering is an unsupervised learning task in which no sample is labeled. For this reason, although thousands of clustering methods have been proposed [1], [2], [3], clustering actually still remains quite challenging [1], [4]. For instance, it is actually not easy for one method to reliably detect the cluster structures in the datasets in Fig. 1.

As one of the classical partitioning-based clustering methods, K-means [9] is regarded as the most popular and widely used clustering method [1], despite that it has some widely known problems. For instance, K-means is sensitive to initialization, requires users to pre-specify the cluster number, and is unable to handle with the non-spherical or unbalanced clusters. Although another popular partitioning-based clustering method, Affinity Propagation (AP) [10], does not require users to specify the cluster number in advance, it requires users to pre-define another non-trivial parameter (i.e., the so-called "preference"), for which an unsuitable setting could lead to the over-partitioning problem. Besides, AP is not as easy and time-saving as K-means, and is not good at detecting non-spherical clusters.

- T. Qiu, K.-F. Yang and Y.-J. Li (Corresponding author) are with the School of Life Science and Technology, University of Electronic Science and Technology of China, Chengdu 610054, China (E-mail: qiutengcool@163.com; yangkf@uestc.edu.cn; liyj@uestc.edu.cn).
- C.-Y. Li is with the School of Life Science and Technology, University of Electronic Science and Technology of China, Chengdu 610054, China, and also with the Center for Life Sciences, Shanghai Institute for Biological Sciences, Chinese Academy of Sciences, Shanghai 200031, China ( E-mail: cyli@sibs.ac.cn).

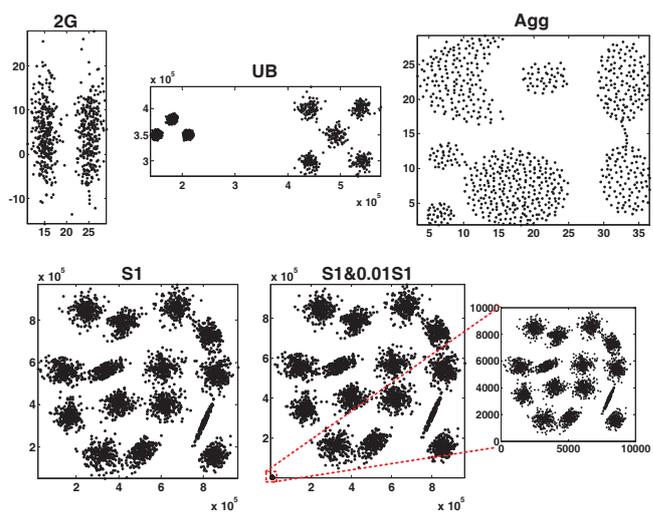

Fig. 1. Five seemingly easy yet challenging synthetical datasets. Although the datasets contain very explicit cluster structures, it is actually not easy for the popular clustering methods to reliably detect them, as detailed in the supplementary material (see Fig. s1). Note that the dataset "S1&0.01S1" consists of two parts: the dataset "S1" and its scaled version with a scaling ratio of 0.01 on each dimension, and thus the clusters from different parts have quite different scales or densities. The datasets "2G" and "S1&0.01S1" are generated by us; the other ones are from literature [5], [6], [7], [8].

Hierarchical clustering (HC) is another classical clustering type, consisting of diverse linkage-based clustering methods (e.g., Single, Complete, Average, etc.) [2]. Since HC clustering methods are generally simple, intuitive, non-parametric, able to graphically show the cluster structure in a hierarchical tree known as Dendrogram, they are widely used especially in biology [11], despite their disadvantages such as the sensitivity to noise and outlier (e.g., for Single linkage) or the cluster shape (e.g., for Complete and Average link-age). Model-based clustering is also a very representative clustering paradigm, in which the most popular method is Gaussian Mixture Model (GMM) [12]. However, GMM has the same open problems as K-means, that is, the sensitivity to initialization and the requirement of setting the cluster number in advance. Spectral clustering such as Normalized cuts (Ncut) [13] and the method proposed by Ng, Jordan and Weiss (N-J-W) [14], and Density-based clustering such as Mean-Shift [15], [16], [17] and DBSCAN [18] are also widely used, due to their advantages in detecting non-spherical clusters. However, Ncut and N-J-W involve demanding spectral decomposition [19], Mean-Shift involves time-consuming iteration [20], [21], [22], [23], and DBSCAN is sensitive to the parameter setting and unable to detect clusters of varying scales or densities [24], [25], [26].

In this paper, we propose a physically inspired graph-theoretical clustering method. We first design a physically inspired rule called **Nearest Descent (ND)**, which makes the dataset organized into an interesting sparse graph called the **In-Tree**. Then the cluster assignment can be simply conducted after removing the undesired edges in this In-Tree graph. The proposed clustering method is simple yet efficient (with a quadratic time complexity) and reliable, able to handle with the datasets with diverse shapes and attributes, and can be flexibly performed in either automatic or interactive (or visualized) ways. In addition, the cluster number is not necessarily required to be specified in advance. The effectiveness of the proposed method is demonstrated by extensive tests on both synthetic and real-world datasets and the comparisons with some widely used methods. This physically inspired graph-theoretical clustering method belongs to the categories of density and hierarchical clustering, but shows significant differences and advantages as compared with the traditional density and hierarchical clustering methods.

This paper is organized as follows: In Section 2, we make detailed descriptions of the proposed method, including an intuitive illustration for its key idea (Section 2.1), basic implementation (Section 2.2), the features and properties of its key elements (i.e., Nearest Descent, in Section 2.3, and the In-Tree, in Section 2.4) and different ways of implementing its certain steps (Section 2.5 and Section 2.6). In Section 3, we give comprehensive analysis about the proposed method, including its relationship with hierarchical clustering (Section 3.1), density clustering (Section 3.2), and gravity clustering (Section 3.3). The experiments are shown in Section 4. The conclusion is provided in Section 5. Some additional figures and texts can be found in the supplementary material.

## 2 METHOD

### 2.1 Overview of the Proposed Method

The proposed clustering method is started from an assumption: what if data points had mass? They would evolve like particles in the physical world. Take the two-dimensional (2D) data points for instance (Fig. 2a) and imagine the 2D space as a horizontally stretched rubber sheet[1]. Intuitively, with the data points put on it, the rubber sheet would be curved downward (Fig. 2b) and the points in the centroids of clusters would have lower potentials. This would in turn trigger the points to move in the *descending direction* of potential and gather at the locations of locally lowest potentials in the end. The cluster members can then be identified.

In this paper, we aim to propose a clustering method by following the above natural evolution and aggregation process of the point system. To this end, we need to answer two basic questions: (i) how to estimate the potential and (ii) how to find the moving trajectories of the points. As for the 1st question, we assume that each point is the center of a local field whose amplitude decreases as the distance to the center increases and that the final potential at each point is estimated as a linear superposition of the fields from all the points. As for the 2nd question, inspired by literature [27], we approximate each trajectory by a zigzag path consisting of a sequence of "short hops" on some transfer points in this point system. For each hop, the nearest point in the descending direction of potential is selected as the transfer point.

Actually, on this relay-like trajectory, we only need to make effort to determine the first transfer point (also called the parent node) for each point. The reason is that once linking each point to its first transfer point by a directed line (also called the edge), one can obtain an attractive network (or graph), called the **In-Tree**, in which the discrete data points are efficiently organized (Fig. 2c). This In-Tree can then serve as a *map*, on which the next hops (or the next transfer points) can be easily determined by following the directions of the edges.[2]

One problem for this In-Tree is that between certain clusters there exist the undesired edges (Fig. 2d) requiring to be removed before the following operation of cluster assignment. Pleasingly, those edges between clusters are quite distinguishable, at least much longer than the edges within clusters. For this reason, one can simply regard the $M$ longest edges as the undesired edges, where $M$ can be determined from the plot in which all the edge lengths are ranked in decreasing order (Fig. 2e). After removing the determined undesired edges, the initial graph will be divided into several unconnected sub-graphs (Fig. 2f), each containing a special node called the root node (corresponding to the point with the lowest potential of one cluster). Within each sub-graph, all the sample points will reach their root node by successively hopping along the edge directions (Fig. 2g and Fig. 2h show two intermediate results[3]). Lastly, the samples that are associated with the same root nodes will be assigned into the same clusters (Fig. 2i).

---

1. It's actually a popular illustration of Einstein's General Relativity, the core of which is summarized, by John Wheeler, as "matter tells space-time how to curve and space-time tells matter how to move". See also http://einstein.stanford.edu/SPACETIME/spacetime2.

2. One can view this graph-based relay (or shift) based on other points of the group as a sort of *Swarm Intelligence*.

3. See all the intermediate results in Fig. s2.

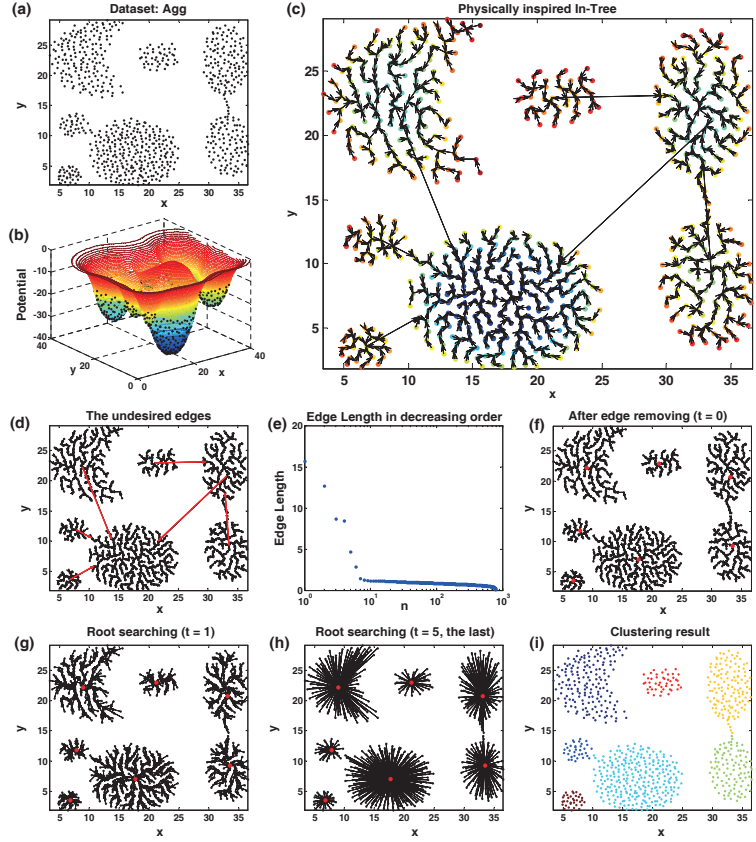

Fig. 2. The idea of the proposed clustering method. (a) The "Agg" dataset consisting of 7 clusters; (b) An illustration for the potential surface. As the isopotential lines vary from red to blue, the potentials in the corresponding areas become lower and lower. (c) The constructed *In-Tree*. The colors on nodes denote the potentials corresponding to the cases in (b). The undesired edges (in red) in the In-Tree is shown in (d). (e) The plot in which the lengthes of all the edges in the In-Tree are arranged in decreasing order. (f) The result after removing the undesired edges. The nodes in red are the root nodes of sub-graphs. (g) The result after the first round of parallel searching of the root node for each non-root node. (h) The results after the last round of parallel searching of the root nodes. All the non-root nodes are directed linked to their root nodes. (i) The clustering result. The samples in the same colors are assigned to the same clusters.

## 2.2 Basic Implementation

Let $\tilde{X} = \{x_i | i = 1, \cdots, N\}$ be a dataset of $N$ sample points and $X = \{i | i = 1, \cdots, N\}$ be the corresponding data indexes. Each point $x_i$ could be either a real-valued or character-typed vector. The proposed method contains the following four stages:

**Stage I, make the dataset organized into the In-Tree structure.** This stage contains several steps:

1) Compute the distance between each pair of points $i, j \in X$:

$$D_{i,j} = d(x_i, x_j). \quad (1)$$

where $d(x_i, x_j)$ denotes certain distance metric (e.g., for Euclidean distance: $d(x_i, x_j) = ||x_i - x_j||_2$).

2) Compute the potential at each point $i \in X$:

$$P_i = -\sum_{j=1}^{N} e^{-\frac{D_{i,j}}{\sigma}}, \quad (2)$$

where $\sigma$ is a parameter, usually called the kernel bandwidth. In Section 2.5, we will show other potential estimation methods, some of which do not require users to set such parameter.

3) Determine all the **"Parent Node Candidates (PNC)"** of each point $i \in X$:

$$J_i = \{j \, | P_j < P_i\} \cup \{j \, | P_j = P_i \text{ and } j < i\}. \quad (3)$$

4) Determine the (true) **Parent Node** of each point $i \in X$:

$$I_i = \begin{cases} \min(\arg\min_{j \in J_i} D_{i,j}), \text{ if } J_i \neq \varnothing \\ i, \qquad\qquad\quad otherwise \end{cases}. \quad (4)$$

Based on the above computations, we can now define a directed graph, denoted as $G = (V, E)$, where $V = X$ denotes the node set and $E = \{\vec{e}(i, I_i) | J_i \neq \varnothing, i \in X\}$ the edge set. For each directed edge $\vec{e}(i, I_i)$, nodes $i$ and $I_i$ are called its start and end nodes, respectively, and its edge length $W_{i,I_i}$ is set as $D_{i,I_i}$. In Section 2.4, we will show this directed Graph $G$ is an In-Tree, in which each point $i$ has only one directed edge started from it (for this reason, we can simply denote $W_{i,I_i}$ as $W_i$), and there is only one node (denoted as $r$) for which $J_r = \varnothing$ and thus there is no directed edge started from this node $r$, as indicated in the definition of the edge set $E$. However, in Eq. 4, we actually set the parent node of the root node $r$ as itself, which is equivalent to adding a *self-loop* on node $r$.[4] Accordingly, we can also define an extended In-Tree $\widetilde{G} = (\widetilde{V}, \widetilde{E})$, where the node set $\widetilde{V} = X$ and the edge set $\widetilde{E} = \{\bar{e}(i, I_i) | i \in X\}$. The following stages are based on this extended In-Tree.

**Stage II, cut the undesired edges.** This stage contains two steps:

1) Determine the undesired edges. Due to the clear difference of the edge lengths between the undesired edges and all the other edges, an intuitive yet effective way is to regard the $M$ longest edges as the undesired edges. One can set the number $M$ in the beginning of the algorithm as what K-means does, or judge from the plot in which the edge lengths $\{W_i | i \in X\}$ are ranked in decreasing order as Fig. 2d shows. We denote the first way as the $M$-**cut** and the second one as the **Edge-plot-based-cut**. In Section 2.6, we will propose other useful methods to determine the undesired edges.

2) Remove the undesired edges. Assume $\omega = \{c_1 \cdots c_M\}$ are the indexes of the start nodes of the $M$ determined undesired edges. One can simply remove the undesired edges by setting $I_i = i$ and $W_i = -\infty$, where $i \in \omega$, i.e., changing the parent node of each node $i \in \omega$ to be itself.

**Stage III, update the parent nodes.** In this stage, the parent node of each node $i$ ($i \in X$) is updated via

$$I_i^{(t+1)} = I_{I_i^{(t)}}^{(t)}, \qquad (5)$$

where $I_i^{(t)}$ denotes the parent node of node $i$ in the $t$-th round of updating, and $I_{I_i^{(t)}}^{(t)}$ the parent node of node $I_i^{(t)}$ (i.e., the parent node of the parent node of node $i$) in the $t$-th round. The updating starts ($t = 0$) after removing the undesired edges and stops in certain round when the parent nodes of all points no longer change, which means each node is directly linked to its root node, as shown in Fig. 2h. It is clear that in each round the updating of the parent nodes for all points can be implemented in parallel.

**Stage IV, assign data points into clusters.** Let $\bar{\omega} = \omega \cup \{r\} = \{c_j | j = 1, \cdots, M + 1\}$, where $c_{M+1} = r$. Given the result in the above stage, the original dataset $X$ can be then divided into $M + 1$ clusters $\Omega = \{C_j | j = 1, \cdots, M + 1\}$, each defined as

$$C_j = \{i | I_i = c_j, i \in X\}, \qquad (6)$$

where each root node $c_j \in \bar{\omega}$ can be viewed as the cluster center.

In comparison, the computation time for the proposed method is mainly consumed in stage I, which makes the proposed method has a time complexity $O(N^2)$. In the supplementary material (Section S4), we provide the complete Matlab code of the proposed method based on $M$-cut, which only involves several lines of codes.

### 2.3 Nearest Descent

If we define PNC simply as $J_i = \{i | P_j < P_i\}$ by ignoring the second term in its original definition of PNC[5] given by Eq. 3, a more explicit understanding is obtained toward the nature of parent node, i.e., ***"for each point, its parent node is the nearest one among the nodes in the descent direction of potential"***. Furthermore, we name this parent node selection method simply as **Nearest Descent (ND)** (in contrast to the classical method of Gradient Descent or Steepest Descent), where

- "Nearest" specifies the local choice (or criterion).
- "Descent" specifies the global direction (i.e., in the descending direction of potential), which serves as a constraint to the local choice.

Thus, ND is actually a "**constrained proximity rule**", different from the classical proximity rules such as $k$-**Nearest-Neighbor** ($k$-NN) and $\epsilon$-**Nearest-Neighbor** ($\epsilon$-NN) in which the neighbors (or the parent nodes) of any node $i$ are simply defined as the $k$ nearest nodes to node $i$ (i.e., $k$-NN) and the nodes within a radius $\epsilon$ centered at node $i$ (i.e., $\epsilon$-NN), respectively. Both $k$-NN and $\epsilon$-NN can be viewed as non-constrained proximity criterions. In addition, ND abandons the term ***Gradient***, which makes it in essence different from the classical gradient-based methods such as **Gradient Ascent (GA)** and **Gradient Descent (GD)**. GA and GD search the local maximum and minimum of a function in the direction of the steepest ascent and descent, respectively, i.e., GA and GD can be viewed as constrained greedy rules, where "constrained" refers to ascent and descent, respectively. Actually, the particularity of ND is well presented via the undesired edges (e.g., Fig. 2d), from which one can find that (i) the parent node of any node $i$ is not necessarily to be its surrounding node within a local sphere, which is unlike $k$-NN or $\epsilon$-NN, and (ii) the node with local minimum of the potential would also shift (or have parent node), which is different from the case in GD.

Besides, it is worth noting that ND itself involves no local operation related parameter (the parameter in the proposed method is involved in the potential computation stage), which is in contrast to $k$-NN and GD (or GA). For $k$-NN, it involves a parameter $k$ to define the number of neighbors. For GA, it involves either an iteration step length parameter, a drawback in the numerical-typed GA-based clustering method called DENCLUE [28], or a parameter to define the local area, a drawback in the graph-theoretical GA-based (GGA) clustering method [29].

Moreover, since ND regards the samples as nodes in Graph Theory, our ND based clustering method could be blind to the underlying features of the samples. Consequently, though inspired from the physical world, ND turns out to be a generalization of the physical world, applicable to the datasets in any high-dimensional space irrespective of whether it is Euclidean space or not. This benefits the proposed clustering algorithm to have wide applications.

---

4. This additional operation can bring simplicity in programming, e.g., there is no need in Stage III to treat the root nodes separately, as revealed by the code in the supplementary material (Section S4). Note that the self-loop will not affect the properties of the In-Tree, i.e., the original graph $G$.

5. The second term aims to (i) guarantee the uniqueness of the parent node and (ii) avoid the unexpected "singleton cluster phenomenon" in some special circumstances. See details in the supplementary material (Section S2).

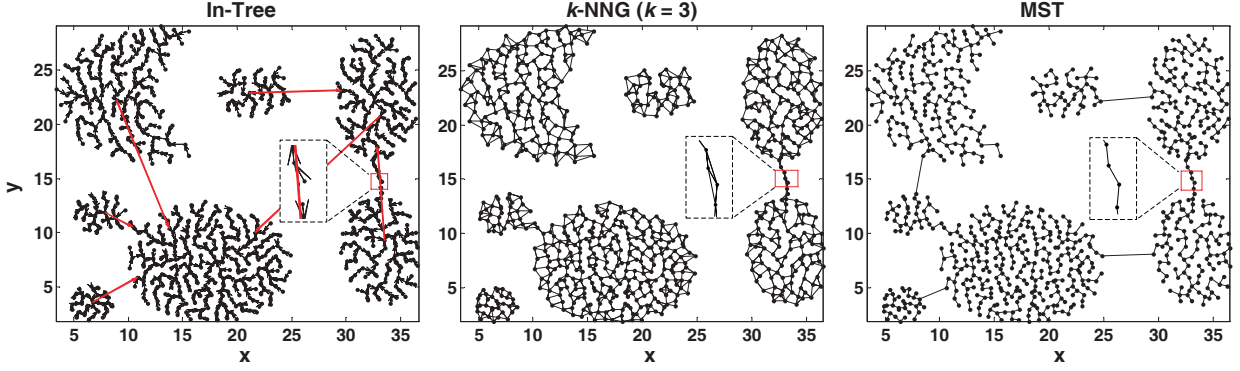

Fig. 3. Comparison between the In-Tree (left), $k$-NNG (middle) and MST (right). Unlike the undesired edge (in red) in In-Tree, both $k$-NNG and MST have very short undesired edges (see the local areas circled by the red boxes) between the close clusters.

### 2.4 In-Tree

**Definition.** In Graph Theory [30], as opposed to the *Out-Tree*, an **In-Tree**, also called in-arborescence or in-branching, is a digraph that meets the following conditions:

(a) Only one node has outdegree 0;
(b) Any other node has outdegree 1;
(c) There is no cycle in it;
(d) It is a connected graph.

**Proof.** It is proved in the supplementary material (Section S3) that the physically inspired graph $G = (V, E)$ constructed in stage-I is an In-Tree defined above.

**Features.** As shown in Fig. 3, compared with other popular graphs like the k-nearest-neighbor graph ($k$-NNG) and Euclidean Minimal Spanning Tree (simply denoted as MST) which are widely used in graph-based clustering [6], [31], [32], [33], [34], [35], [36], [37], [38], [39], the physically inspired In-Tree constructed by ND shows two significant features which make it a better candidate for clustering.

1) Like MST and $k$-NNG, this physically inspired In-Tree well follows the cluster structure in the dataset.
2) Although there are some undesired edges between certain clusters, those edges are quite distinguishable, in contrast to the cases in MST and $k$-NNG.

The 1st feature guarantees the meaningfulness of the discovered clusters. The 2nd feature makes it possible to easily and reliably determine those undesired edges by diverse methods.

**Properties.** Besides the above specialties of this physically inspired In-Tree, the In-Tree graph structure itself has very good properties [30], which facilitate the reliability and efficiency of clustering:

1) Removing any edge in In-Tree will divide the In-Tree into two unconnected sub-graphs, each still being an In-Tree.
2) There is one and only one directed path between each non-root node and the root node.

The 1st property guarantees that after removing $M$ undesired edges, there will be $M + 1$ unconnected sub-graphs and thus the original dataset $X$ will be divided into $M + 1$ clusters $\{C_1, \cdots, C_{M+1}\}$, with $C_i \neq \varnothing, i \in \{1, \cdots, M + 1\}$; $\bigcup_{i=1}^{M+1} C_i = X$; $C_i \cap C_j = \varnothing, i, j \in \{1, \cdots, M + 1\}, i \neq j$.

The 2nd property guarantees that when searching[6] along the directed edges on the generated sub-graphs, each non-root node is sure to reach its root node in the same sub-graph in finite steps.

### 2.5 Methods to Compute Potential

The potential estimation in Eq. 2 (which involves the only parameter of the proposed method) is close to the kernel density estimation in statistics [40], except the difference in sign. A more general relationship between potential $P$ and density $\rho$ is that, the denser the area is, the lower the potential in that area will be. In other words, the above relationship can be written as $P = f(\rho)$, where $f(\rho)$ is a monotonically decreasing function such as $f(\rho) \propto -\rho$ or $f(\rho) \propto 1/\rho$. For this reason, potential estimation here can benefit from the massive researches in density estimation in statistics [40], [41], [42], [43]. Since the proposed method cares the relative potentials at sample points, one can flexibly transfer the density into potential, by choosing appropriate decreasing functions.

For instance, in literature [41], the Gaussian Kernel Density Estimation via diffusion (KDEd[7]) is provided in which the parameter (i.e., the Gaussian kernel bandwidth $\sigma$) can be estimated automatically. In this case, once obtaining the density estimation, one can simply set the potential as $P_i = -\rho_i$.

In literature [42], theoretical analysis reveals that the density at any point $x_i$ can be simply estimated based on the k-nearest-neighbor rule, that is,

$$\rho_i = \frac{(k-1)}{N \cdot V_\xi(x_i)} \tag{7}$$

where $\xi$ denotes the distance between $x_i$ and its $k$-th nearest neighbor, $V_\xi(x_i)$ denotes the volume of the region within the radius $\xi$ centered at $x_i$, and the optimal $k$ was estimated in literature [44]. In this case, one can simply set the potential as $P_i = V_\xi(x_i)$.

Besides from the field of statistics, we can also benefit from other fields like astronomy [45], [46] which also involves the density estimation problem. For instance, in literature [45], non-parametric density estimation methods are provided based on Delaunay Triangulation (DT) [8] or its dual Voronoi tessellation. The corresponding methods are called Delaunay tessellation field estimation (DTFE) or Voronoi tessellation field estimation (VTFE), respectively. DTFE defines the density at each node $i$ as $\rho_i \propto 1/V(\Gamma_i)$, where $\Gamma_i$ denotes the neighboring nodes of node $i$ and $V(\Gamma_i)$ denotes the volume of the regions formed by these neighboring nodes. Accordingly, we can derive two simpler yet still efficient ways for potential estimation. The first one (denoted as **DT-I**) is written as $P_i = V(\Gamma_i)$. Another one (denoted as **DT-II**) is to approximate the potential by some statistical features (e.g., mean) of the distances between node $i$ with its neighboring nodes $V(\Gamma_i)$. For instance, the mean-based potential is defined as

$$P_i = \frac{1}{|\Gamma_i|} \sum_{j \in \Gamma_i} d_{i,j}, \qquad (8)$$

where $|\Gamma_i|$ denotes the number of nodes in $\Gamma_i$.

---

6. This search on the graph is in essence different from the numerical iteration in the methods such as Mean-Shift.
7. http://web.maths.unsw.edu.au/~zdravkobotev/.

### 2.6 Methods to Determine the Undesired Edges

In the 2nd Stage of 2.2, the undesired edges are removed using the simple $M$-cut or Edge-Plot-based-cut. The main problem for these two methods is that they cannot deal with the dataset (e.g., the dataset "S1&0.01S1" in Fig. 1) containing the clusters of largely varying scales or densities. Due to the close relationship of the proposed method with hierarchical and density-based clustering methods (see Sections 3.1 and 3.2), in the following we propose some visualization-based edge removing methods, which can solve the above problem to a great extent.

**Scatterplot-based cutting**. Each directed edge, $\vec{e}(i, I_i)$, $i \in X$ in the extended In-Tree $\vec{G}$ is actually associated with as many as three features: the edge length $(W_i)$, the potential of its start node $P_i$, and the potential of the end node $P_{I_i}$. Unlike the feature $W_i$, either $P_i$ or $P_{I_i}$ alone cannot make the undesired edges distinguishable. Nonetheless, inspired by the *Decision Graph* proposed by Rodriguez and Laio [47], the potential can be used as an auxiliary feature of the edge length. Specifically, one can represent each directed edge by two features such as $W_i$ and $P_i$. Then, all the edges can be displayed in a 2D scatterplot (Fig. 4a, Type I). Since only the undesired edges have large values for both $W_i$ and $P_i$, the undesired edges will pop out in the scatterplot and thus can be interactively determined easily. Actually, similar results can be obtained if each directed edge is represented by $W_i$ and $P_{I_i}$ (Type II), or by $W_i$ and $(P_i + P_{I_i})/2$ (Type III). See details in the supplementary material (Fig. s4).

**Dendrogram-based cutting**. The motivation for this method is to transform the In-Tree into the *Dendrogram*, a 2D diagram widely used in Hierarchical Clustering. The transformation can be simply implemented as follows: the In-Tree is first transformed to an undirected tree structure. This means to construct a sparse distance matrix $S$, with the non-zero elements $S_{i,I_i} = S_{I_i,i} = W_i$, $i \in X$. Then, given the sparse matrix $S$, a Dendrogram can be obtained by running Single Linkage hierarchical clustering. Unlike the above Scatterplot which only provides a visualization of the key features of the In-Tree, the Dendrogram provides a more complete visualization of the In-Tree, from which users can perceive the cluster structure and get access to the information of the samples associated with the leaf nodes non-overlapped at the bottom of the dendrogram. Also, the dendrogram can help define a linear order of the rows and columns of the color-coded data table known as heat map [11], [48], [49] so that similar rows or columns are placed closely and the clusters can be easily perceived in the re-ordered heat map. All of these are very useful for biologists, for example, in the massive high-throughput gene expression cluster analysis [11]. Then, an appropriate threshold can be set to cut the Dendrogram or the corresponding In-Tree (Fig. 4d). In addition, this Dendrogram-based method can benefit from fruitful studies on Dendrogram, such as the methods to perform the local cuts[9] on the dendrogram either automatically [50], [51] or interactively [52], [53], [54], and the methods to re-order the leaf nodes in the Dendrogram for better visualization of the clusters [55], [56], [57].

**Barplot-based cutting**. Inspired by the so-called *reachability plot* (actually a 2D bar plot, as detailed in Section 3.2) in OPTICS [58] and the connection between the Dendrogram and the reachability plot [59], we can transform the extended In-Tree into a 2D reachability-plot-like barplot in which all the directed edges in the extended In-Tree are vertically (with the start nodes[10] downward and end nodes upward) arranged in a linear order. The arrows of the directed edges can be omitted, whereas the edge lengths are preserved in the bar plot. The key question is how to determine the linear order. A simple way is to take the order of the leaf nodes in the above Dendrogram as the linear order of the bottom nodes in the bar plot[11]. The relatively longer bars in the constructed bar plot (Fig. 4e) correspond to the undesired edges in the In-Tree and thus they can be interactively determined one by one. Actually, in practice, one can simply pick out the top nodes of the corresponding bars as an equivalence to picking out the undesired edges. This barplot-based cutting has two advantages: (i) the bar plot is advantageous for displaying the dataset with large size, as compared with the Dendrogram [59]; (ii) Compared with the local cuts in Dendrogram, it is much easier to implement the process of interactively determining the undesired edges one by one (which can also be viewed as the local cuts) in the bar plot.

**Data Representations and Edge-cutting Methods.** Actually, the In-Tree (Fig. 2c), the edge-length-based plot (EP) (Fig. 2e), the Decision-Graph-like Scatterplots (SP) (Fig. 4 a), the dendrogram (DE) (Fig. 4b), and the barplot (BP) (Fig. 4c) can be viewed as five different data representations of the original datasets, and the last four data representations (i.e., EP, SP, DE and BP) are derived from the In-Tree (see an illustration in Fig. s6). Unlike the In-Tree[12], all the derived data representations can be always visualized in a 2D space (irrespective of the dimension of the test dataset), from which the flat partitioning can be obtained interactively. Compared with EP, the other three derived data representations (i.e., SP, DE and BP) are more effective ones[13], from which the flat clustering results could be more reliably obtained, especially in handling with the dataset consisting of the clusters with varying scales or densities. Nevertheless, due to the distinguishable difference of the undesired edges in the In-Tree (as revealed by the comparison between the In-Tree, MST and $k$-NNG in Fig. 3), the simple EP-based cutting could actually deal with many practical problems (as proved by the Experiments in Section 4).

---

8. See details in the supplementary material (Fig. s3).

9. This can solve the problem that simply setting a global threshold in the dendrogram may fail to detect the clusters of largely varying densities.

10. Note that the start nodes correspond to the samples in the dataset.

11. Similar to OPTICS, one can also obtain the linear order via a walk on the In-Tree.

12. Actually, besides the derived data representations, the undesired edges can be also reliably removed directly based on the In-Tree in an interactive or semi-supervised way if some additional requirements can be met. See details in the supplementary material (Section S5).

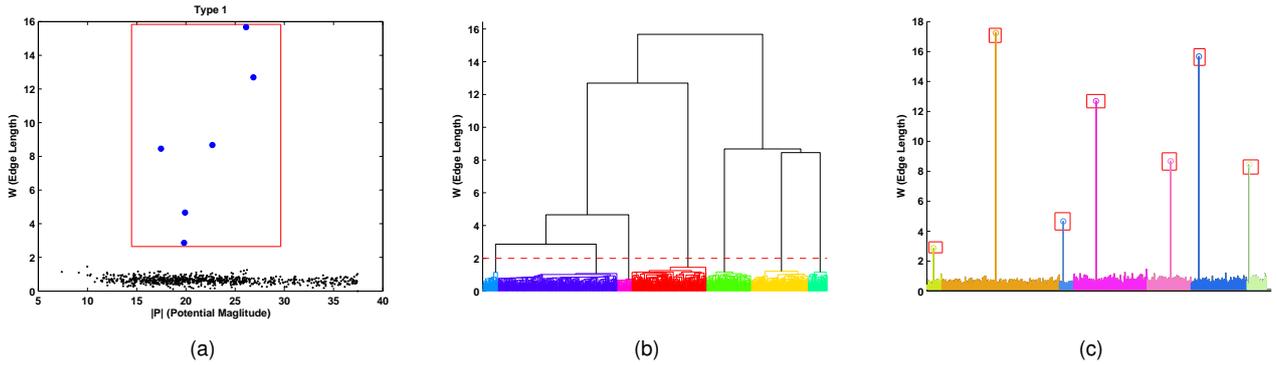

Fig. 4. Illustration for different methods of removing the undesired edges. (a) Scatterplot-based cutting (Type I). The pop-out points within the red rectangles correspond to the undesired edges. (b) Dendrogram-based cutting. The red dashed line denotes the global threshold for cutting the Dendrogram. (c) Barplot-based cutting. The bars with their top nodes circled by the red rectangles correspond to the undesired edges. All the cutting methods in (a ∼ c) can lead to the same clustering result as shown in Fig. 2. Note that in the dendrogram shown in (b) we can also interactively cut the branches one by one (see the supplementary material: Fig. s5), which is more flexible than the threshold-based global cutting; in the bar plot shown in (c) we associate the root node with an edge length, being the maximum edge length on the In-Tree plus a very small constant, in order to make the cluster containing the initial root node distinguishable.

## 3 ANALYSIS

### 3.1 General Hierarchical Clustering

Clustering algorithms can be divided into two broad categories: partitional clustering (PC) and hierarchical clustering (HC) [1], [2]. PC algorithms (such as K-means) directly seek for a flat partitioning based on some optimization criteria. HC algorithms, as defined in one of the most popular review article [1], "recursively find nested clusters either in agglomerative mode (starting with each data point in its own cluster and merging the most similar pair of clusters successively to form a cluster hierarchy) or in divisive (top-down) mode (starting with all the data points in one cluster and recursively dividing each cluster into smaller clusters)".[14]

Comparatively speaking, it would be more appropriate to classify the proposed method into the category of HC. However, in contrast to HC, the proposed method nests data points neither in the above agglomerative nor divisive mode. Instead, it adds link on each point in a parallel way, through which a tree structure can be constructed in only one layer. To some degree, this shows the limitation in the traditional definition of the hierarchical clustering. To overcome this limitation, here we make a definition for a new category, the *General Hierarchical Clustering (G-HC)*.

**Definition.** *General Hierarchical Clustering seeks from the dataset for an efficient data representation, based on which the optimal flat partitioning of the dataset could be easily obtained.*

Compared with the traditional HC, G-HC gives a more general definition for hierarchical clustering and thus could have more general meaning. For G-HC, a hierarchical process of organizing dataset is not necessarily needed, nor is the Dendrogram. In other words, the traditional HC is a special case of G-HC.

**Bu-Td framework.** According to the definition, G-HC actually contains two phases: the *bottom-up (Bu)* and *top-down (Td)* phases. Bu phase aims to obtain an efficient data representation. This data representation could be a special data structure, such as a Dendrogram (e.g., traditional Linkage-based HC), a tree (e.g., the proposed method and the MST-based clustering [31]), a $k$-NNG (e.g., Chameleon [32]), a Shared Neareat Neighbor graph (e.g., SNN clustering [24]), a condensed Dendrogram (e.g., HDBSCAN [60]), or simply a reordering of the dataset (e.g., OPTICS [58]). This data representation could also be a special feature space, such as a feature-based scatterplot [47], or the spectral space (e.g., spectral clustering methods Ncut [13] and N-J-W [14]). One can even view the estimated probability density function such as in GMM as another special data representation[15]. Td phase seeks to extract an optimal partitioning result from the above data representations.

Obviously, the proposed method contains the above two phases of G-HC and thus can be viewed as a hierarchical clustering method in terms of the general definition of the hierarchical clustering. This is also the case for other non-traditional HC methods listed above e.g., MST-based clustering, Chameleon, SNN, HDBSCAN, OPTICS, etc. [16]

---

13. Compared with the transformation from In-Tree to EP, the transformations from In-Tree to the data representations such as the SP, DE and BP preserve more information, which means less information loss or distortion during the transformations.

14. As the definition shows, HC contains two opposite branches: the Agglomerative HC (AHC) and Divisive HC (DHC). Compared with DHC, AHC is more popular. In the following, HC refers to AHC.

15. For GMM, based on the mixture probability density function, a flat partitioning can be easily obtained using the Bayesian criterion.

**Advantages of the Bu-Td framework for the proposed method.** (i) Theoretically, since the methods belonging to G-HC share the same framework (i.e., Bu-Td), the proposed method could benefit from other methods in the category of G-HC. For instance, in the Bu phase, one data representation could be transformed to another data representation to facilitate a more reliable flat partitioning in the Td stage, as revealed in Section 2.6. In fact, similar transformation can be also found in the recent method HDBSCAN: the transformation from MST to Dendrogram, and then to the condensed Dendrogram. (ii) The Bu-Td framework endows the proposed method with the capability of resisting the "density noise", which also helps increase the reliability of the proposed method. This will be detailed in Section 3.2.

### 3.2 Comparison with Density Clustering

Density clustering is built on the following assumptions [26], [61]: (i) the data points are sampled from a density function; (ii) the data points belonging to different clusters are located in high density areas separated by relatively low density ones. To extract the above density-based clusters, many methods have been proposed. Here, we divide them into two major categories: *the density-peak-based (1st category)* clustering methods (e.g., DENCLUE [28], Mean-Shift[17] [15], [16], [17], GGA [29] and Decision Graph [47]) and *the density-border-based (2nd category)* clustering methods (e.g., DBSCAN [18], OPTICS [58] and HDBSCAN [60]). See details in the supplementary material (Section S1). The 1st category can be further subdivided into the *gradient-based* (e.g., DENCLUE, Mean-Shift and GGA) and *non-gradient-based* (e.g., Decision Graph) methods.

The relationship between the proposed method and density clustering is quite obvious, due to the close relationship between the potential and density as shown in Section 2.5. In the density perspective, the parent node selection rule, i.e., the method of Nearest Descent, can be viewed as *the method of Nearest Ascent*. Like Nearest Descent, here Nearest Ascent means that each point jumps to the nearest point in the ascending direction of density, or simply speaking, each point ascends to the nearest point. Thus, the proposed method can be viewed as *a non-gradient density-peak-based clustering method using Nearest Ascent*.

**Comparison with the gradient-based methods in the 1st category.** The density-peak-based clustering based on the gradient ascent[18] faces the following problems of (i) introducing an extra parameter in searching the density peaks (e.g., for DENCLUE and GGA), (ii) involving time-consuming numerical iteration (e.g., for Mean-Shift), (iii) being only applicable to the real-valued dataset (e.g., for DENCLUE and Mean-Shift), and (iv) being susceptible to the convergence (e.g., for DENCLUE and GGA); (v) being sensitive to density estimation or the kernel bandwidth (e.g., for DENCLUE, Mean-Shift and GGA).

---

16. Notice that although Chameleon, OPTICS and HDBSCAN are usually called the hierarchical methods, here we contribute to making it clear that at which hierarchical clustering they strictly belong to.

17. Actually, the idea of Mean-Shift can be also found in literature [62], [63].

18. Specifically, *DENCLUE* combines the classic gradient ascent with a special data structure, which solves the 2nd problem. *Mean-Shift* derives a variant form of the gradient ascent, which solves the 1st problem. And the 3rd problem could be partly solved if the kernel function is carefully selected. *GGA* defines a graph-theoretical-based gradient ascent, which solves the 2nd and 3rd problems. See details in the supplementary material (Section S1).

However, the proposed method can avoid all the above problems: it involves neither the step-length-like parameter (in the case of a given density estimation) nor the limitation for data attributes, due to its non-gradient and graph-theoretical characteristics as revealed in Section 2.3; it involves neither numerical iteration nor the problem of convergence, due to the (2nd) property of the In-Tree as revealed in Section 2.4; it is insensitive to the density estimation especially for the under-estimated cases due to the Bu-Td framework (Section 3.1), the reason of which is explained as follows. When the density is under-estimated, there would generate some noise-like bumps (or spurious density peaks) on the density surfaces of certain clusters, which would lead to the over-partitioning results for methods such as DENCLUE and GGA, since each bump would be regarded as a cluster center. However, the Bu-Td framework of the proposed method could contribute to avoiding that: in the Bu phase, instead of being regarded as the root node, each density-peak (DP) node (irrespective of being spurious or genuine) is associated with a parent node, which generates a directed edge between them. However, compared with the edges (usually between clusters) generated by the genuine DP nodes, the edges generated by the spurious DP nodes are usually the ones within the clusters and thus less distinguishable and then in the Td phase less likely to be determined as the undesired edges and removed, which prevents the spurious DP nodes from becoming the cluster centers at last.

**Comparison with the non-gradient-based methods in the 1st category.** In 2014, Rodriguez and Laio (RL) [47] proposed a non-gradient density clustering method via fast determining the density peaks from a feature extraction perspective, that is, when each point $i$ is featured by such two variables as the local density $\rho_i$ and the smallest distance $\delta_i$ between point $i$ and the other points with higher densities, the density peaks are obviously the only ones with relatively high values in both variables (as compared to the non-density-peaks points) and thus will pop out as outliers in the 2D scatterplot (i.e., the so-called *Decision Graph* where all data points are plotted based on the values of the two variables above. The key for RL's method or the Decision Graph is the method of defining the distance variable $\delta_i$, which is in essence the same as Nearest Ascent. Specifically, the distance variable $\delta_i$ can be viewed as the length of the edge between point $i$ and its parent node $I_i$ defined by Nearest Ascent. For this reason, the proposed method and RL's method can be viewed as the two methods which arrive at the same position from two different routes.

Compared with RL's method, the contribution of this paper are as follows: (i) we reveal the nature of the common idea, i.e., Nearest Ascent (or Nearest Descent), in the perspective of physics, which helps to make a direct connection and significant comparison with the classic method such as gradient ascent (Section 2.3) and gradient-ascent-based density clustering methods (Section 3.2), through which the difference and contributions of the idea could be more sufficiently presented. (ii) We introduce for the first time the graph structure, In-Tree, into the clustering field. (iii) We present a more general framework (i.e., estimate the density, construct the In-Tree and remove the undesired edges in the In-Tree), from which one will find (in Section 2.6) that RL's method can be viewed as a special case (i.e., the scatterplot-based cutting, Type I) of the methods in determining the undesired edges, and one will also find that Nearest Descent is the simplest method but not the most time-saving method in constructing the In-Tree [64], [65]. (iv) We reveal the relationship of the proposed method with hierarchical clustering and further generalize the traditional hierarchical clustering to a larger class, i.e., G-HC.

**Comparison with the methods in the 2nd category.** We can find very similar development for the two categories: both from non-G-HC to G-HC (see definition in Section 3.1). Specifically, for the 1st category, this refers to the development from DENCLUE to ND or RL's method; for the 2nd category, from DBSCAN to HDBSCAN or OPTICS. This surprising consistency makes it possible that the proposed method (1st category) could benefit from the density clustering methods in the 2nd category, as partly revealed in Section 2.6.

### 3.3 Comparison with Gravity clustering

Gravity clustering can be traced back to the literature [66] in 1977. Since then, many gravitational methods were proposed [67], [68], [69], [70], [71]. Like the proposed method, most gravitational clustering methods assume that (i) the data points have masses, alike the particles in physical world and (ii) the particle system would undergo a process of natural dynamic evolution, through which the data point would gradually get clustered at last and then the members in each cluster could be identified. However, unlike the proposed method, most previous methods are based on the equations of Newtonian's law of gravity and Newtonian's second Law of motion, such as the algorithms in literature [67], [68], [69], [70]. Newtonian's law of gravity describes the universal attractive force between any two particles $i$ and $j$, written as $F \propto m_i m_j / d_{i,j}$, where $m_i$ and $m_j$ are the masses of the two particles and $d_{i,j}$ denotes the Euclidean distance between the two particles. Note that for a particle in the particle system, it will be attracted by all the other particles and thus the resultant force should be considered. Newtonian's second Law of motion reveals how a particle will move if a force is exerted on it, which can be symbolically written as $F = ma$, where $m$ is the mass of the particle, $a$ the acceleration, and $F$ the force. Based on the acceleration, the displacement, $s$, can be determined due to the relationship $a = d^2(s)/dt^2$. Besides, there are also methods such as the two gravitational algorithms in literature [71], which use the non-dynamic ways to cluster the data points based on features of the local resultant forces.

A common problem for the Newtonian's force-based methods, whether being dynamic or non-dynamic, is that they require a non-trivial number of parameters, as high as 4 in literature [67], [68], [71]. In contrast, the number of parameter in the proposed method is no more than 1.

Note that although Ruta and Gabrys [72] proposes a dynamic clustering field model built on the concept of the field similar to the proposed method, it explicitly uses the concept of force again when considering the movement of each data point and the force is defined based on the gradient of the field. Thus, the clustering method in literature [72] is very similar to the gradient-based density clustering methods as introduced previously and thus will also suffer from almost all the problems that gradient-based density clustering methods have (see details in Section 3.2).

## 4 EXPERIMENTS

### 4.1 The test datasets

Table 1 lists the features of all the test datasets.

**The synthetical datasets.** The first five datasets are shown in Fig. 1, all with very clear cluster structures, which could guarantee the validity of the clustering evaluation. The 6th dataset "D512" is a 512-dimensional dataset containing 16 Gaussian-distributed clusters [6].

**The real-world datasets[19].** "*Wine*": measurements of chemical features (e.g., Alcohol, Malic acid, etc.) of wines derived from three different cultivars. "*Seeds*": measurements of geometrical features (e.g., area, perimeter, etc.) of the internal kernel structures presented in the X-ray images of three varieties of wheat. "*Cancer*" (full name: "Breast Cancer Wisconsin"): measurements of geometrical features (e.g., radius, area, etc.) of the cell nuclei in the digitized images of fine needle aspirate of breast mass, either Benign or malignant. "*Iris*": measurements of geometrical features (e.g., petal length, petal width, etc.) of iris plants derived from three different cultivars. "*Banknote*": measurements of the statistical features (e.g., variance, curtosis, etc.) of wavelet transformed images taken from genuine and forged banknote specimens. "*Mushroom*": measurements of external characteristics (stalk-shape, odor, etc.) of 23 species of gilled mushrooms, each labeled as edible or poisonous.

### 4.2 Compared methods

As listed in Tables 3 and 4, we compared the proposed method, denoted as CND (i.e., Clustering by Nearest Descent), with some widely used methods of different types: Hierarchical clustering (HC), Partitioning-based Clustering (PC), Model-based clustering (MC), Spectral clustering (SC), Density Clustering (DC).

### 4.3 Experimental details

**Distance metric.** For the real-valued datasets, the simple Euclidean distance was used. For the character-typed dataset (i.e., the "Mushroom"), the pair-wise distance between samples $x_i$ and $x_j$ was defined as the number of features that are different. The same distance metric was used for different methods when testing the same datasets.

**Pre-processing.** The datasets marked with asterisk in Tables 3 and 4 denote they were pre-processed (including either the dimensionality reduction or normalization). Specifically, for the datasets "D512", "Iris" and "Seeds", each feature was normalized to the unit range (i.e., between 0 and 1). For the dataset "Wine", it was first embedded into a low dimensional space using the principal component

TABLE 1
The Test Datasets Employed in This Work.

| | | Synthetic | | | | | Real-World | | | | | |
|---|---|---|---|---|---|---|---|---|---|---|---|---|
| Dateset Index | | 1 | 2 | 3 | 4 | 5 | 6 | 7 | 8 | 9 | 10 | 11 | 12 |
| Dateset Name | | UB | 2G | Agg | S1 | S1&0.01S1 | D512 | Seeds | Wine | Cancer | Iris | Banknote | Mushroom |
| Dataset Details | $N$ | 6500 | 600 | 788 | 5000 | 10000 | 1024 | 210 | 178 | 699 | 150 | 1372 | 8124 |
| | $\ell$ | 2 | 2 | 2 | 2 | 2 | 512 | 7 | 13 | 9 | 4 | 4 | 22 |
| | $m$ | 8 | 2 | 7 | 15 | 30 | 16 | 3 | 3 | 2 | 3 | 2 | 2 |
| | Attr. | r. | r. | r. | r. | r. | r. | r. | r. | r. | r. | r. | char. |

$N$: the number of sample points; $\ell$: the number of features; $m$: the number of classes in the ground truth. Attr.: the attributes of each feature (r.: real-valued, char.: character-typed).

analysis (PCA) [20] and then each feature was normalized to the unit range.

**Evaluation indexes**. We used two widely used indexes [73]: the **Accuracy (AC)** and the **Normalized Mutual Information (NMI)**. Both indexes range from 0 to 1, indicating how well the cluster assignments match with the true labels. A higher value represents better performance.

**Cluster number**. For each dataset, we tried to guarantee for each compared method that $|\Omega| = m$, where $|\Omega|$ denotes the cluster number obtained by a clustering method and $m$ is the class number in the ground truth. But for the tests of HDBSCAN on the datasets "AGG", "Wine" and "Iris", we could not make $|\Omega| = m$ regardless of how the parameter was valued. In those circumstances, we chose the cluster numbers that are most close to the class numbers.

**HC methods**. For HC, we chose the most popular linkage methods (e.g., Single, average, etc.). For each method, an appropriate threshold was set to cut the corresponding dendrogram into $m$ clusters for each dataset.

**K-means and GMM**. For both K-means and GMM, we reported the best results out of 20 random initializations. Note that K-means and GMM cannot deal with the non-numerical datasets like "Mushroom".

**DBSCAN and HDBSCAN**. For both methods, we tuned the parameters (two for DBSCAN, one for HDBSCAN) carefully to report the best results as possible as we can, and we assigned the data points that are regarded as noise to the nearest clusters. Note that although HDBSCAN can handle the dataset "Mushroom", the available R package "dbscan"[21] for HDBSCAN has not yet provided the function of supporting the dissimilarity matrix as the input.

**SC methods**. Following the suggestion in literature [74], we used the Gaussian similarity function (involving a kernel bandwidth parameter $\sigma$) to transform the distance to similarity measurement and construct the similarity graph based on the $k$-nearest neighbor graph, which is believed to be able to make the SC methods become less sensitive to parameter setting. Thus, in the experiments, SC methods (i.e., Ncut and N-J-W) involve two parameters ($\sigma$ and $k$). We chose two empirical values for them as reference: $k = 10$, and $\sigma = \sigma_0 = (\sum_{i,j} d_{ij})/N^2$ (i.e., the average of all of the distances between all pairs of samples). For the SC methods, since they need to run K-means in the end, we also ran K-means 20 times for them, and reported the best results.

20. The dimension of the data after dimensionality reduction is 6, with 86% energy being preserved.

21. https://cran.r-project.org/web/packages/dbscan/index.html

**The proposed method CND**. (i) *Potential estimation*: as shown in Table 2, we tried to use the DT- and KDEd-based potential estimation methods defined in Section 2.5, in order to reduce the manual effort in tuning the parameter for achieving good performance.[22] DT-based methods (i.e., DT-I and DT-II) free the proposed method from any parameter. KDEd can estimate the kernel parameter $\sigma$ automatically. Only for the last dataset (i.e., "Mushroom" dataset), we estimated the potential using Eq. 2 and set the bandwidth $\sigma$ as the empirical value $\sigma_0$. (ii) *The method to determine the undesired edges*: for the dataset "S1&0.01S1", we used the Scatterplot-based cutting (actually, other cutting methods in Section 2.6 can be also applied to this dataset), while for all the other datasets, we simply regarded the $m - 1$ longest edges as the undesired edges (i.e., using the $M$-cut).[23]

### 4.4 Experimental results

The results on the synthetical and real-world datasets are shown in Table 3 and Table 4, respectively. We can see that it is actually not easy for any clustering method to achieve ideal results on all the datasets. Nevertheless, overall, the proposed method (CND) performs better as compared with other widely used clustering methods. Specifically, on the 1st, 2nd, 4th, 5th, 6th and 10th datasets, the proposed method obtains the scores of both evaluation indexes AC and NMI either equivalent or very close to 1 (i.e., the highest values for both indexes), showing almost ideal performance and ranking the first or joint first in comparison to other methods. On the 3rd, 7th, 8th and 9th datasets, although the proposed method does not perform best, the scores it achieves are still quite acceptable (AC > 0.9) and comparable with the best results achieved by other methods. On the rest two datasets "Banknote" and "Mushroom", however, the proposed method achieves relatively poor

22. Generally, KDEd and DT-based potential estimation methods are comparable. In Table 2, we show the methods that could report the relatively better performance. However, note that KDEd was unable to handle with the 5th dataset ("S1&0.01S1") well. Besides, since the 12th dataset ("Mushroom") is not a real-valued dataset, DT-based methods and KDEd cannot be directly applied on it.

23. Due to the saliency of the undesired edges, $M$-cut can automatically determine the undesired edges of the In-Trees for most datasets, whereas $M$-cut would encounter problem when handling with the multi-scale datasets such as "S1&0.01S1". Specifically, $M$-cut cannot determine the undesired edges between the small-scale clusters in "S1&0.01S1", since those edges are even shorter than the non-undesired edges within the large-scale clusters. Compared with $M$-cut, Scatterplot-based cutting is more powerful and flexible in handling with such type of dataset (as detailed in Fig. s7).

TABLE 2
The Potential Estimation (P. E.) Methods Used for Different Datasets

| Dataset Index | 1 | 2 | 3 | 4 | 5 | 6 | 7 | 8 | 9 | 10 | 11 | 12 |
|---|---|---|---|---|---|---|---|---|---|---|---|---|
| P. E. Method | KDEd | KDEd | KDEd | KDEd | DT-II | KDEd | DT-I | DT-I | KDEd | DT-II | DT-II | Eq. 2 ($\sigma = \sigma_0$) |

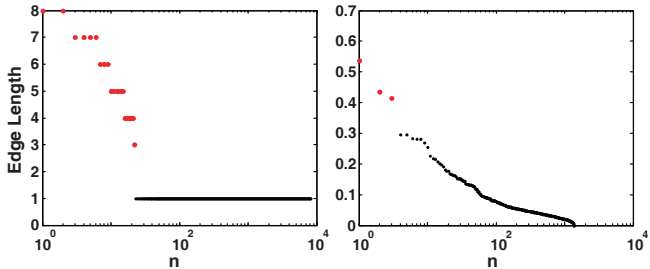

Fig. 5. The edge-length-based plots of the datasets "Mushroom" (left) and "Banknote" (right). The plots indicate that for "Mushroom" and "Banknote", 22 and 3 undesired edges (i.e., the red dots) should be removed, resulting in 23 and 4 clusters, respectively, which are much more than the class numbers of the two datasets (see Table 1).

scores, despite ranking respectively the first and the fifth in comparison to other methods. The problem may lie in the fact that for any of the two datasets, the proper cluster number might not be in line with the class number provided in the ground truth, as revealed in the edge-length-based plots (Fig. 5). Actually, the proposed method can also achieve almost ideal performances ("Mushroom": purity[24] = 1; "Banknote": purity = 0.991) on the two datasets, when the cluster numbers ("Mushroom": 23; "Banknote": 4) are determined based on the edge-length-based plots in Fig. 5.

In Section 3.2, we have analyzed the advantage of the proposed method over the representative density-peak-based methods. Here, in Tables 3 and 4, we can also find the advantage of the proposed method over the representative density-border-based clustering method such as DB-SCAN. DBSCAN performs poor in dealing with the dataset "S1&0.01S1", which verifies the well-known problem of DBSCAN in detecting the clusters with varying densities [24], [25], [26]. Compared with DBSCAN, although its recent variant HDBSCAN successfully solves this problem as proved by its performance on the dataset "S1&0.01S1", HDBSCAN performs relatively poor on the datasets "AGG", "Seeds", "Wine" and "Iris", which could be due to the relatively poor density estimation for the density-border-based methods as compared with the density-peak-based methods. See a detailed analysis in the supplementary material (Section S1).

Besides, compared with DBSCAN, the proposed method also has apparent advantage in parameter setting. Specifically, for one thing, the proposed method can utilize special potential estimation methods (e.g., DT-I, DT-II or KDEd) to free users from the task of manually setting the parame-ter. For another, even the parameter is set manually, the proposed method is not as sensitive to the parameter as DBSCAN. Take the dataset "S1&0.01S1" for instance. The proposed method can achieve almost ideal clustering results when the kernel parameter $\sigma$ varies drastically (from 100 to 1000), as shown in Fig. 6.

## 5 CONCLUSION

In this paper, we proposed a physically inspired clustering method, which first makes the dataset organized into the In-Tree structure and then focuses on removing the undesired edges. The high saliency of the undesired edges guarantees that they can be easily determined by various edge-removing methods. In turn, the diversity and effectiveness of those edge removing methods also prove the effectiveness of the proposed method.

Overall, the proposed method has the following features: (i) it is inspired by Einstein's global perspective of the relationship between mass and movement, in contrast to the pair-wise gravitational force defined by Newtonian's law of gravity; (ii) it mimics the dynamic clustering process of the particle system based on the Graph Theory rather than the numerical iteration; (iii) it is a non-gradient density-peak-based clustering method, theoretically superior to the widely used gradient-ascent- (or descent-) based density clustering paradigm; (iv) it uses a hierarchical rather than a flat-partitioning clustering paradigm and, strictly speaking, belongs to the newly-defined General Hierarchical Clustering; (v) it for the first time introduces the In-Tree data structure into clustering, which shows significant advantages as compared with MST and $k$-NNG in terms of the features of the undesired edges between clusters; (vi) it allows the generating of the undesired edges (or "errors") in the early phase, which contributes to its simplicity, flexibility, and the reliability to resist other uncertain challenges such as in handing with the clusters of varying scales or densities and the problem of the poorly estimated potential or density; (vii) it supports diverse visualization and interaction, which could help increase the reliability of the results [75].

In summary, the proposed method is a physically inspired graph-theoretical density-based (*General*) hierarchical clustering method, using a non-gradient paradigm (i.e., Nearest Descent) and a particular graph structure (i.e., In-Tree). The proposed method is simple yet efficient and reliable, and is applicable to various datasets with diverse shapes, attributes and any high dimensionality.

## ACKNOWLEDGMENTS

This work was supported by the Major State Basic Research Program (2013CB329401), the Natural Science Foundations of China (61375115, 91420105), the 111 Project (B12027). The authors would like to thank Hong-mei Yan, Ke Chen, Ye-zhou Wang and Shao-bing Gao for the suggestions, Fu-ya

24. Note that the codes for AC and NMI provided by literature [73] require that the cluster number should not be larger than the class number. Thus, the purity was used here, which is another popular external evaluation index [4], [7], ranging from 0 to 1. A higher value represents better clustering performance. Besides, for the "Banknote" dataset in this test, each feature was normalized to the unit range.

TABLE 3
The Performance of Different Methods on the Synthetic Datasets

| Methods | | 1, UB | | 2, 2G | | 3, AGG | | 4, S1 | | 5, S1&0.01S1 | | 6, D512* | |
|---|---|---|---|---|---|---|---|---|---|---|---|---|---|
| | | AC | NMI | AC | NMI | AC | NMI | AC | NMI | AC | NMI | AC | NMI |
| HC | Single | 0.984 | 0.985 | 0.502 | 0.002 | 0.824 | 0.801 | 0.475 | 0.659 | 0.305 | 0.488 | 1.000 | 1.000 |
| | Complete | 0.686 | 0.697 | 0.533 | 0.007 | 0.793 | 0.863 | 0.986 | 0.977 | 0.459 | 0.597 | 1.000 | 1.000 |
| | Average | 1.000 | 1.000 | 0.503 | 0.001 | 1.000 | 1.000 | 0.991 | 0.983 | 0.524 | 0.596 | 1.000 | 1.000 |
| | Centroid | 1.000 | 0.999 | 0.510 | 0.006 | 0.996 | 0.989 | 0.991 | 0.983 | 0.526 | 0.595 | 1.000 | 1.000 |
| | Median | 1.000 | 1.000 | 0.537 | 0.004 | 0.980 | 0.955 | 0.885 | 0.919 | 0.441 | 0.595 | 1.000 | 1.000 |
| | Weighted | 0.692 | 0.697 | 0.550 | 0.009 | 0.845 | 0.817 | 0.796 | 0.906 | 0.426 | 0.592 | 1.000 | 1.000 |
| | Ward | 1.000 | 0.999 | 0.673 | 0.090 | 0.838 | 0.880 | 0.992 | 0.985 | 0.434 | 0.597 | 1.000 | 1.000 |
| PC | K-means | 0.608 | 0.705 | 0.998 | 0.984 | 0.786 | 0.835 | 0.994 | 0.986 | 0.864 | 0.947 | 0.873 | 0.926 |
| MC | GMM | 0.835 | 0.816 | 0.998 | 0.984 | 0.734 | 0.793 | 0.926 | 0.954 | 0.511 | 0.621 | 0.816 | 0.934 |
| SC | Ncut | 0.838 | 0.867 | 0.998 | 0.984 | 0.819 | 0.867 | 0.900 | 0.947 | 0.832 | 0.906 | 1.000 | 1.000 |
| | N-J-W | 0.788 | 0.804 | 0.998 | 0.984 | 0.956 | 0.922 | 0.925 | 0.951 | 0.915 | 0.961 | 1.000 | 1.000 |
| DC | DBSCAN | 1.000 | 1.000 | 0.998 | 0.984 | 0.994 | 0.981 | 0.994 | 0.987 | 0.543 | 0.797 | 1.000 | 1.000 |
| | HDBSCAN | 1.000 | 0.999 | 0.998 | 0.984 | 0.827 | 0.803 | 0.992 | 0.983 | 0.995 | 0.992 | 1.000 | 1.000 |
| | **CND** | 1.000 | 1.000 | 0.998 | 0.984 | 0.994 | 0.981 | 0.995 | 0.990 | 0.994 | 0.990 | 1.000 | 1.000 |

The datasets marked with "*" are the ones with preprocessing.

TABLE 4
The Performance of Different Methods on the Real-World Datasets

| Methods | | 7, Seeds* | | 8, Wine* | | 9, Cancer | | 10, Iris* | | 11, Banknote | | 12, Mushroom | |
|---|---|---|---|---|---|---|---|---|---|---|---|---|---|
| | | AC | NMI | AC | NMI | AC | NMI | AC | NMI | AC | NMI | AC | NMI |
| HC | Single | 0.348 | 0.011 | 0.382 | 0.014 | 0.657 | 0.002 | 0.660 | 0.579 | 0.555 | 0.001 | 0.522 | 0.005 |
| | Complete | 0.843 | 0.609 | 0.657 | 0.309 | 0.797 | 0.252 | 0.880 | 0.744 | 0.647 | 0.116 | 0.714 | 0.144 |
| | Average | 0.895 | 0.702 | 0.399 | 0.023 | 0.940 | 0.649 | 0.887 | 0.770 | 0.663 | 0.092 | 0.523 | 0.006 |
| | Centroid | 0.919 | 0.735 | 0.404 | 0.018 | 0.969 | 0.786 | 0.660 | 0.579 | 0.663 | 0.092 | 0.519 | 0.001 |
| | Median | 0.605 | 0.295 | 0.393 | 0.028 | 0.654 | 0.001 | 0.880 | 0.744 | 0.644 | 0.054 | 0.506 | 0.023 |
| | Weighted | 0.814 | 0.556 | 0.382 | 0.038 | 0.938 | 0.648 | 0.880 | 0.744 | 0.663 | 0.098 | 0.506 | 0.023 |
| | Ward | 0.871 | 0.683 | 0.904 | 0.717 | 0.963 | 0.762 | 0.887 | 0.770 | 0.536 | 0.004 | 0.890 | 0.580 |
| PC | K-means | 0.886 | 0.664 | 0.972 | 0.890 | 0.959 | 0.732 | 0.887 | 0.736 | 0.612 | 0.029 | – | – |
| MC | GMM | 0.943 | 0.825 | 0.882 | 0.651 | 0.873 | 0.533 | 0.967 | 0.898 | 0.532 | 0.005 | – | – |
| SC | Ncut | 0.857 | 0.649 | 0.978 | 0.909 | 0.632 | 0.015 | 0.893 | 0.761 | 0.570 | 0.017 | 0.551 | 0.069 |
| | N-J-W | 0.886 | 0.677 | 0.972 | 0.894 | 0.658 | 0.003 | 0.900 | 0.770 | 0.585 | 0.035 | 0.519 | 0.000 |
| DC | DBSCAN | 0.624 | 0.468 | 0.753 | 0.471 | 0.731 | 0.101 | 0.680 | 0.588 | 0.577 | 0.026 | 0.506 | 0.023 |
| | HDBSCAN | 0.552 | 0.468 | 0.646 | 0.453 | 0.928 | 0.608 | 0.667 | 0.584 | 0.576 | 0.024 | – | – |
| | **CND** | 0.919 | 0.741 | 0.961 | 0.849 | 0.959 | 0.733 | 0.967 | 0.885 | 0.742 | 0.328 | 0.522 | 0.005 |

The datasets marked with "*" are the ones with preprocessing.

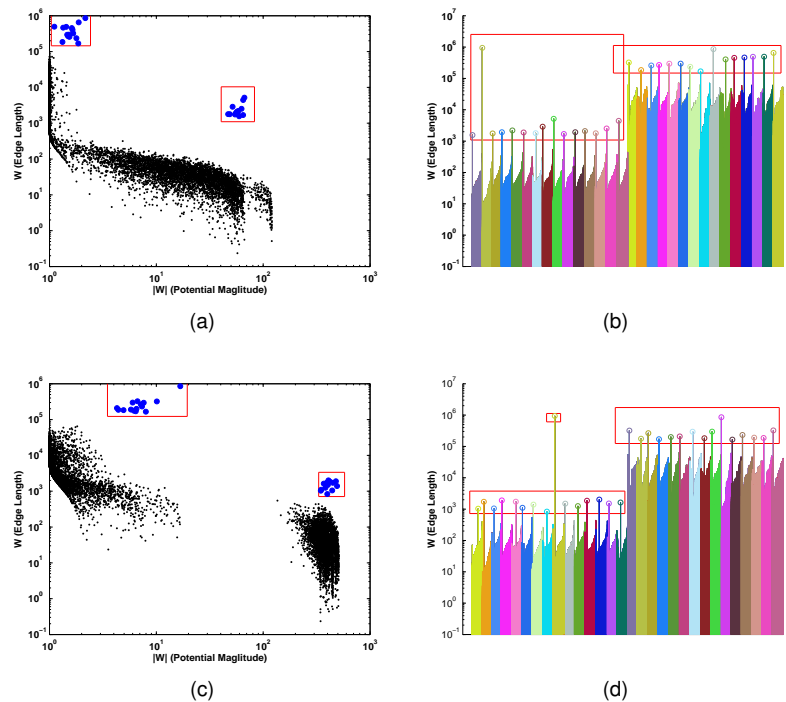

Fig. 6. The tests on the dataset "S1&0.01S1" with the parameter varying widely. When $\sigma = 100$, the proposed method can achieve an almost perfect performance (AC = 0.993, NMI = 0.988) using either the scatterplot-(a) or barplot-based cutting (b). When $\sigma = 1000$, the proposed method can also achieve an almost perfect performance (AC = 0.991, NMI = 0.987) using either the scatterplot- (c) or barplot-based cutting (d).

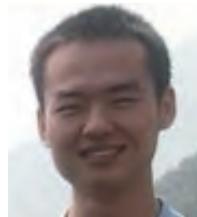

**Teng Qiu** received M.Sc. degree in Biomedical engineering from University of Electronic Science and Technology of China (UESTC) in 2015, where he is currently pursuing his Ph.D. degree. His research interest is clustering.

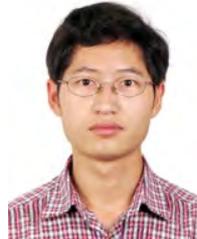

**Kai-Fu Yang** received the Ph.D. degree in biomedical engineering from University of Electronic Science and Technology of China (UESTC), Chengdu, China, in 2016. He is an Associate Professor with the Key Laboratory for Neuroinformation, Ministry of Education, School of Life Science and Technology, UESTC. His research interests include visual mechanism modeling, computer vision, and cognitive computing.

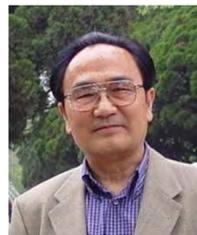

**Chao-Yi Li** graduated from the Chinese Medical University in 1956 and Fudan University in 1961. He became an academician of the Chinese Academy of Sciences in 1999. He is currently a professor of UESTC, and professor of Shanghai Institutes for Biological Sciences. His research interest is visual neurophysiology.

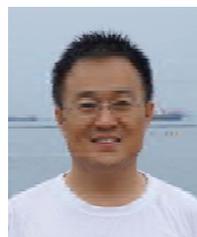

**Yong-Jie Li** (M'14–SM'17) received the Ph.D. degree in biomedical engineering from University of Electronic Science and Technology of China (UESTC), Chengdu, China, in 2004. He is a Professor with the Key Laboratory for Neuroinformation, Ministry of Education, School of Life Science and Technology, UESTC. His research interests include visual mechanism modeling, image processing, and intelligent computation.

Supplementary Material to

# Nearest Descent, In-Tree, and Clustering

Teng Qiu, Kai-Fu Yang, Chao-Yi Li, and Yong-Jie Li

This supplementary material contains:

Figures: s1 $\sim$ s10

Sections: S1 $\sim$ S5

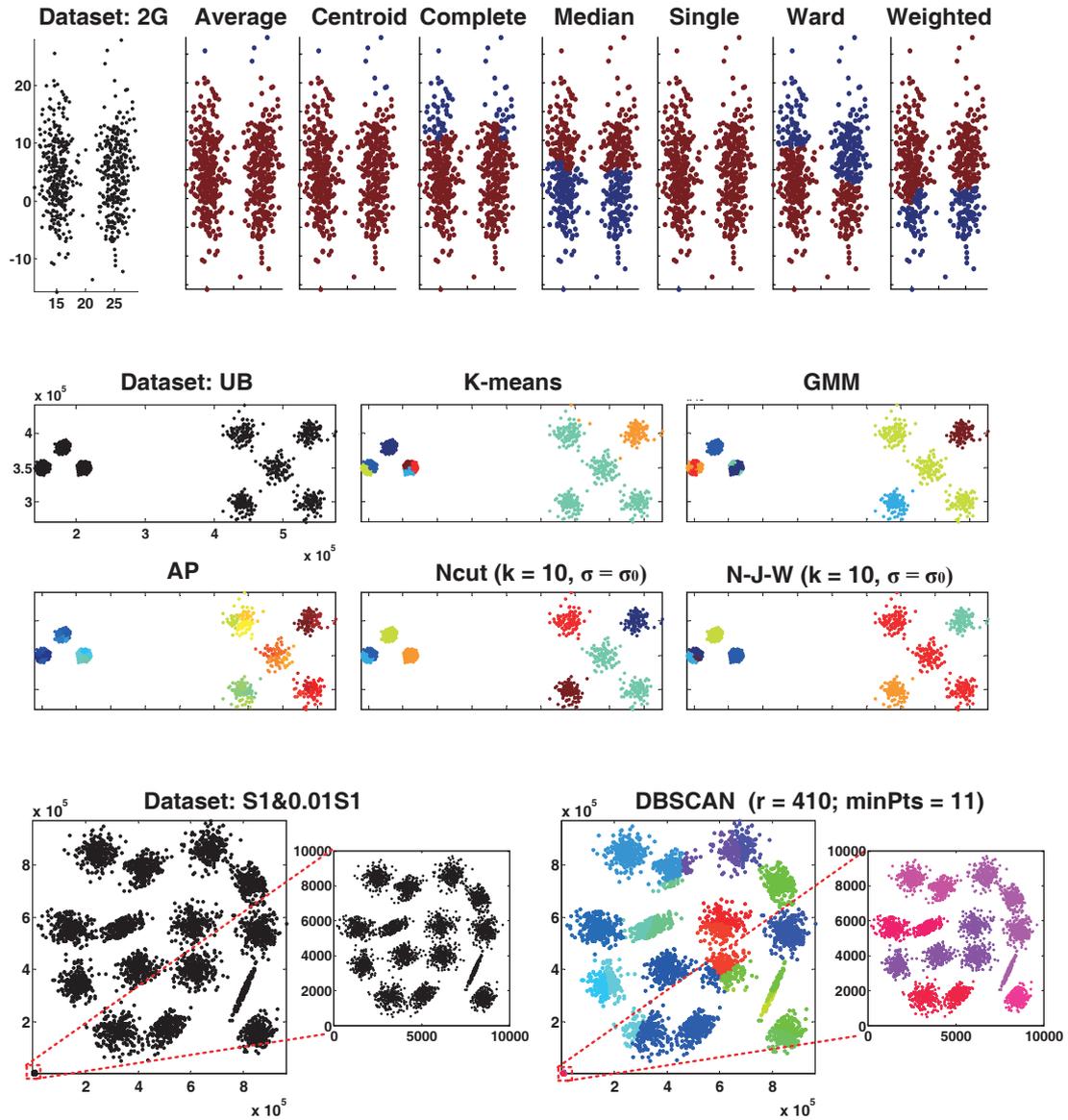

Fig. s1. The results of some popular methods on several synthetical 2D datasets. This figure shows the typical false clustering results of the popular methods, including the Hierarchical clustering methods (average, centroid, complete, median, single, ward and weighted linkage), K-means, GMM, AP, Ncut, N-J-W and DBSCAN. $\sigma_0$ denotes the average of all of the distances between all pairs of samples. The small-scale clusters in the dataset S1&0.01S1 are zoomed in.

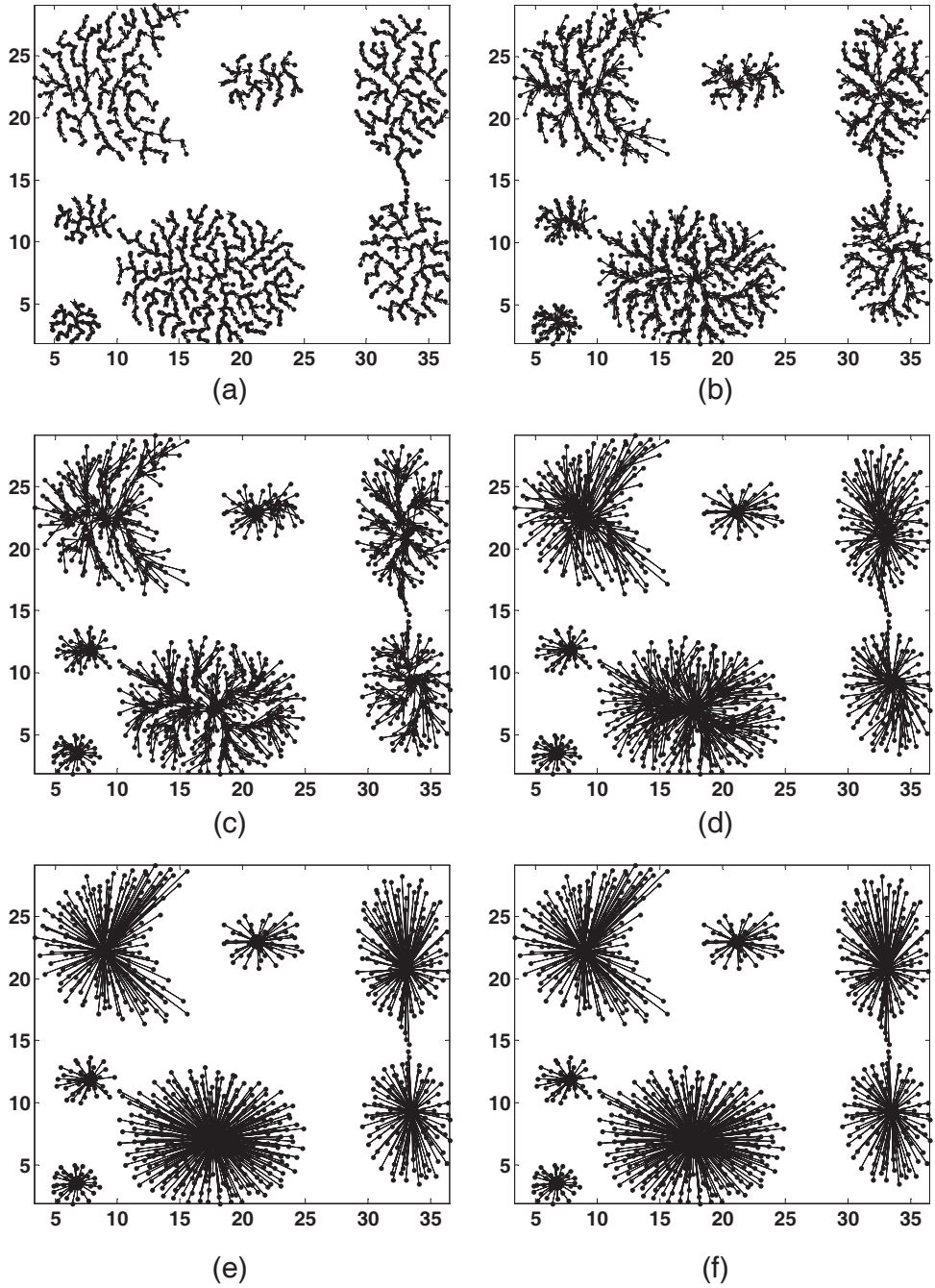

Fig. s2. The whole process for searching the root nodes in the proposed method. (a) The result after removing the undesired edges. (b ~ f) The results after the 1st to the 5th parallel search, respectively.

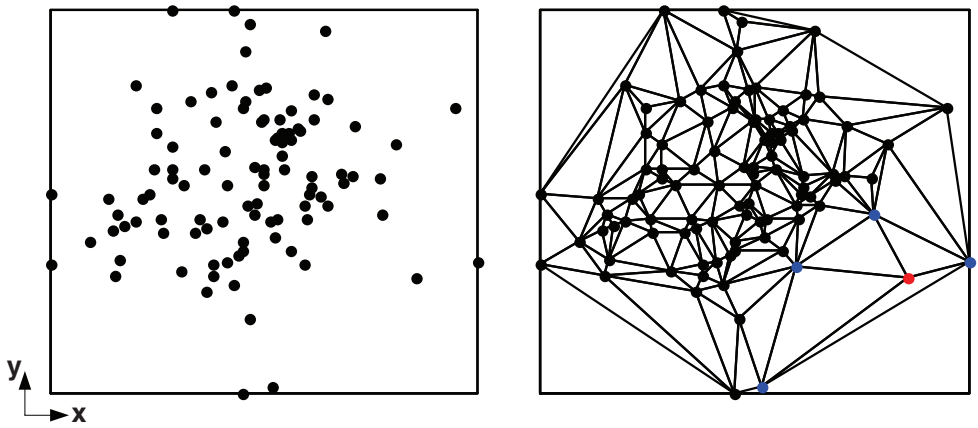

Fig. s3. An illustration for the 2D Delaunay Triangulation. Left: the input dataset. Right: the generated Delaunay Triangulation. For the red point, the nodes in blue are its neighbor nodes. Here the 2D Delaunay Triangulation is a set of triangulations such that no point is inside the circumcircle of any triangle.

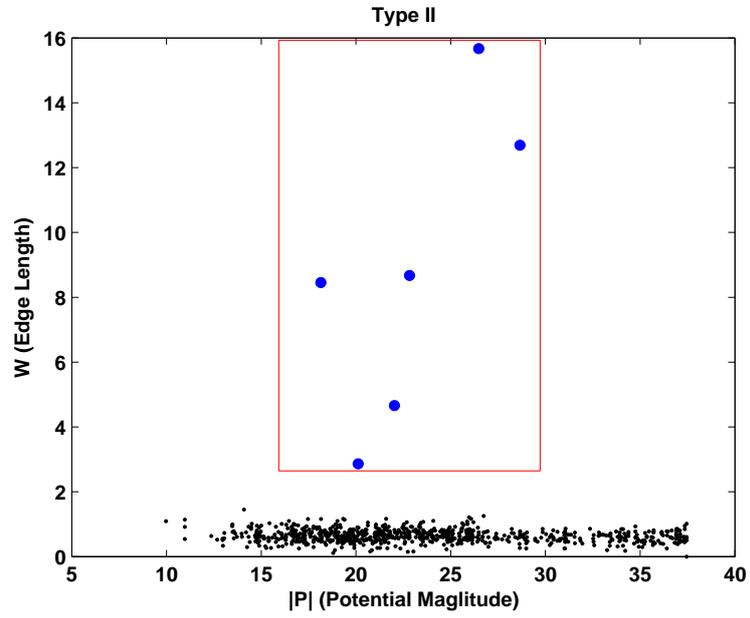

(a)

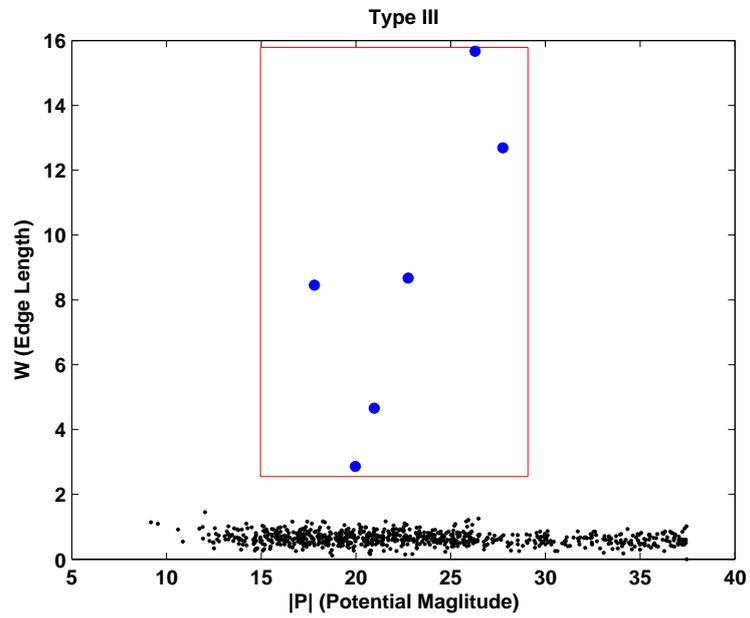

(b)

Fig. s4. Two types of scatterplot-based edge cutting methods. (a) Type II. (b) Type III. The scatterplots of the Types II and III show very similar patterns to Type I (Fig. 4a), indicating that these three types (Types I, II and III) are actually comparable in practice.

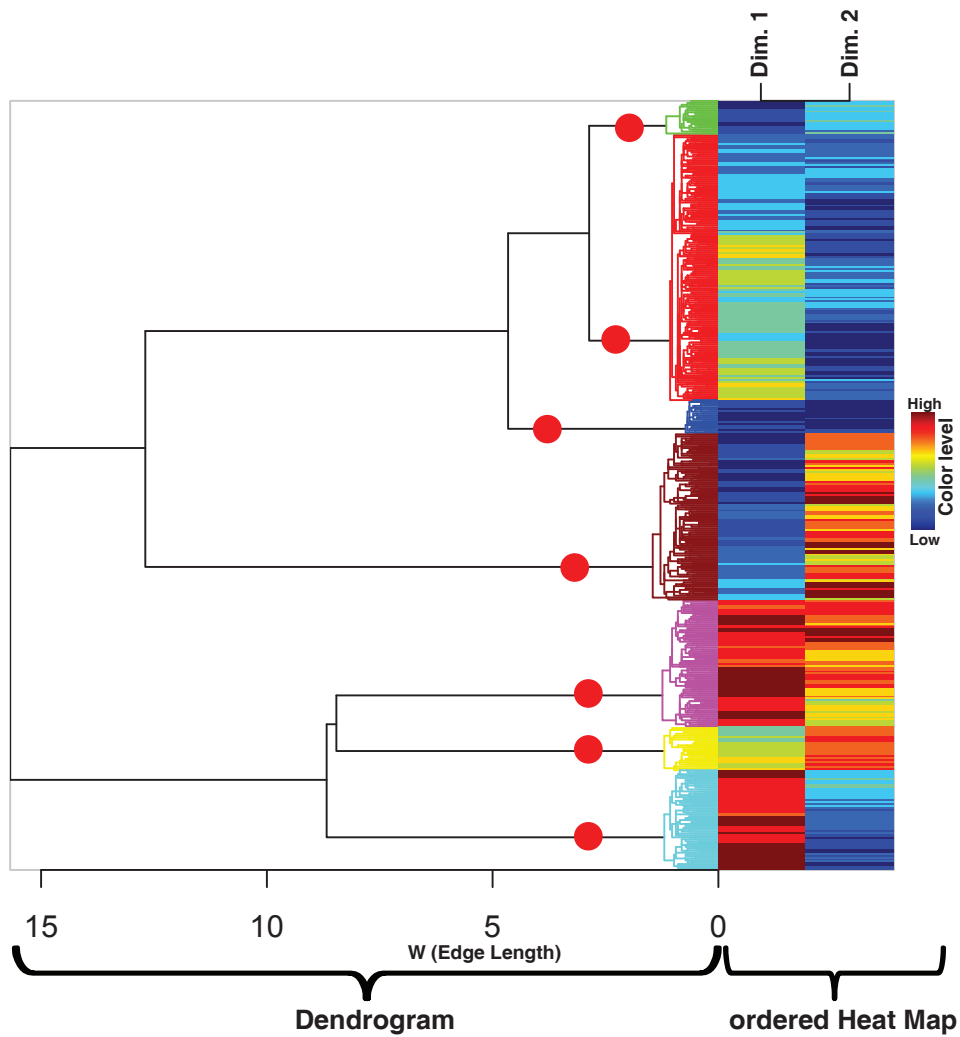

Fig. s5. Illustration for interactively cutting the branches on the In-Tree-based Dendrogram one by one, based on the R package "idendr0" in literature [1]. The branches associated with red dots are the ones that are interactively determined. In the right side, the heat map of the test data ("Agg") is arranged according to the order of the leaf nodes of the Dendrogram. Note that in the heat map, rows represent samples and columns the dimensions (theoretically, the samples could be of any high dimension). The values in the data matrix are displayed according to the color level, through which one can perceive the dissimilarity of two samples in the dataset easily. This (arranged) heat map clearly shows that the samples within clusters have smaller distances than those between clusters, which reveals the reliability of the clustering result.

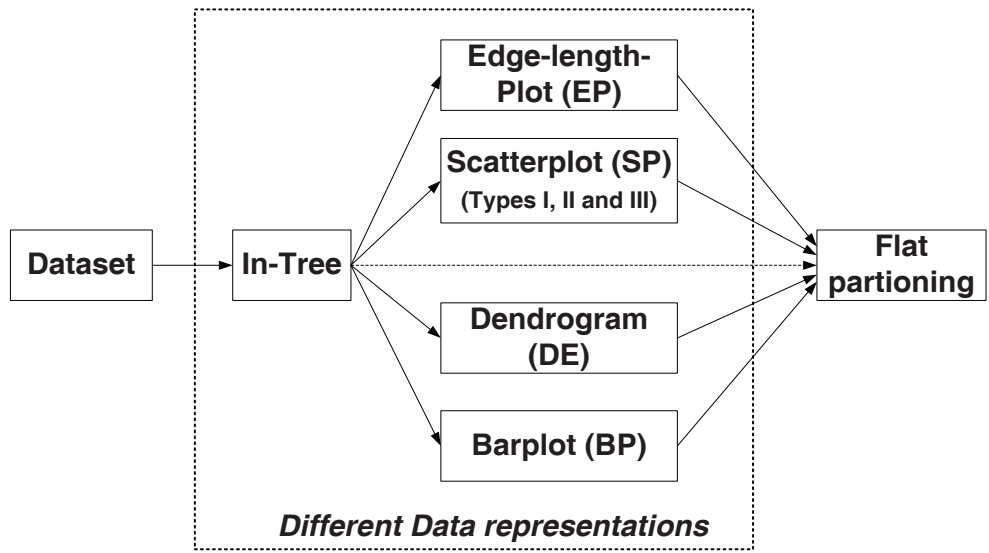

**(a)**

1) ***Whether supports the interactive cutting:***
• In-Tree: support for the 1D or 2D dataset;
• EP, SP, DE and BP: support for any high dimensional dataset.

2) ***Whether supports the local cutting***
   ***(in contrast to using a global threshold):***
• In-Tree: support for the 1D or 2D dataset; support for any high dimensional dataset if there are few labeled data (i.e., in the semi-supervised learning case).
• SP, DE and BP: support.
• EP: not support.

**(b)**

Fig. s6. An illustration for showing the relationship (a) and features (b) of different data representations. The data representations EP, SP, DE and BP are derived from the initial data representation, i.e., the In-Tree.

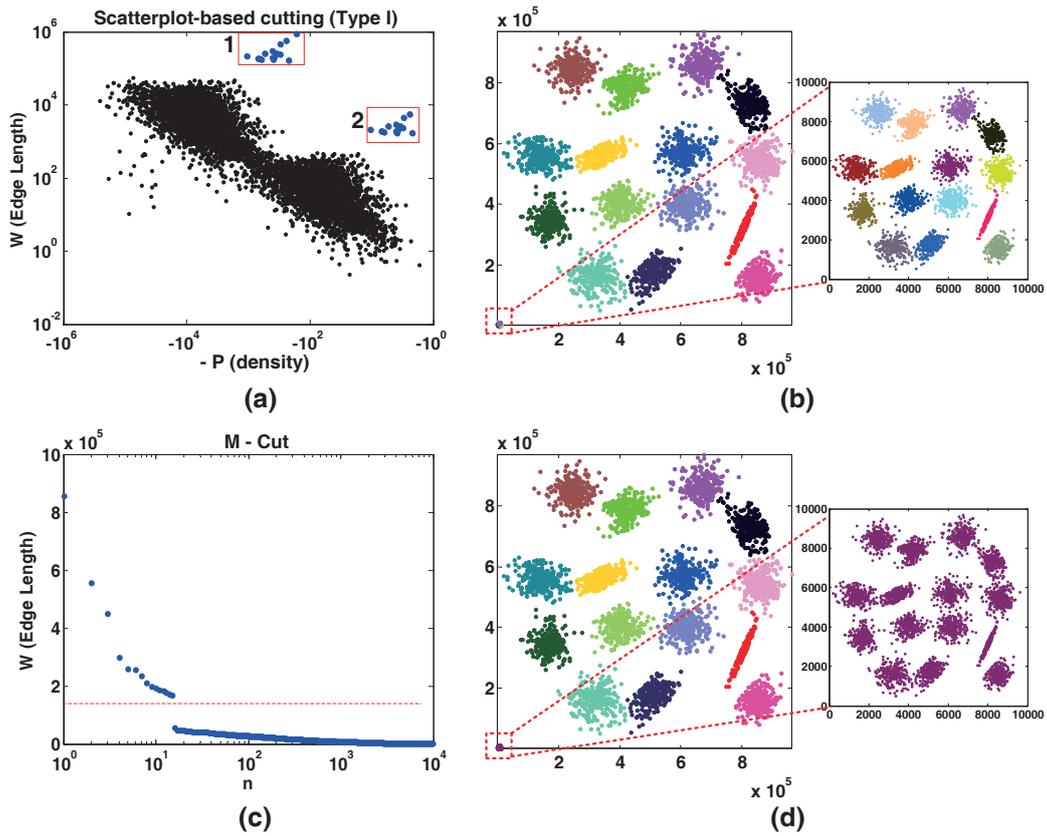

Fig. s7. The tests on the dataset "S1&0.01S1" using scatterplot-based cutting (a) and $M$-cut (c) with the corresponding clustering results shown in (b) and (d), respectively. The scatterplot in (a) contains two groups of pop-out points (i.e., the blue dots in the red boxes). In the experiment, we interactively determined them one by one. The points in the 1st and 2nd groups actually correspond to the undesired edges between the large-scale clusters and the ones between the small-scale clusters, respectively, as proven in (b), in which the 30 clusters (the small-scale clusters are zoomed in) are almost perfectly found. Obviously, if we use $M$-cut, which plots the edge lengths in decreasing order as shown in (c), those undesired edges between the small-scale clusters will be mixed with other non-undesired edges within the large-scale clusters and thus it is impossible to determine them, as proven in (d), in which only 16 clusters are found with all the small-scale clusters being regarded as one cluster. Note that in (c), the red dashed line denotes the global threshold. The edges with the lengths larger than the threshold will be regarded as the undesired edges.

## S1 DENSITY-BASED CLUSTERING

Without loss of generality, let's consider a given density function of a continuous argument $x$ with its domain denoted as the set $\Psi$. Then density-based clustering can be simply viewed as a task of separating the set $\Psi$. For **the 1st category (density-peak-based)**, a cluster is defined as a maximal connected sub-set of the data points that would converge at the same density peak if they are forced to climb on the surface of the density function. For **the 2nd category (density-border-based)**, a cluster is defined as a maximal connected sub-set of the dataset $\Psi$ for which the corresponding densities are greater than a threshold. Note that in practice the input for clustering is a dataset $\tilde{X} = \{x_i | i = 1, \cdots, N\}$ (which can be viewed as the domain of a discrete augment) where each point $x_i$ could be any high-dimensional vector and the attribute of $x_i$ could be diverse (e.g., numerical, character-typed, etc.).

**Representative methods in the 1st category: DEN-CLUE, Mean-Shift, GGA and RL's Decision Graph.** The most intuitive and straight-forward idea in this category is to first estimate an explicit density function $f(x)$ underlying in the dataset $X$ via the kernel density estimation, and then let each point $i$ climb (or ascend) along the direction of the *gradient* of the density function using the following numerical iteration

$$x(n+1) = x(n) + \lambda \cdot \frac{\nabla f(x(n))}{||\nabla f(x(n))||},$$

where $\lambda$ defines the shifting step length, $x(n)$ is the current position, $x(n+1)$ is the next position, $\nabla f(x(n))$ denotes the gradient at the position $x(n)$ and $|| \cdot ||$ computes the Euclidean norm of a vector in the Euclidean space. This iteration process starts at $x(0) = x_i$ and is expected to stop at certain optimum of the density function $f(x)$ where the gradient is zero. At last, the points belonging to the same clusters will converge at the same density peaks. This gradient-ascent-based (or steepest-ascent-based) idea can be actually found in the literature as early as in 1975 [2]. Despite being theoretically reasonable, clustering via the above gradient ascent has at least the following implicit problems in practice: (i) sensitive to the parameter in the iteration, i.e., the step length; (ii) involving time-consuming iteration;[1] (iii) only applicable to numerical (real-valued) datasets;[2] (iv) not always guaranteed to converge;[3] (v) sensitive to the result of density estimation (or the kernel bandwidth).[4] The popular method DENCLUE [3] solves the 2nd problem by combining the above gradient ascent and an efficient data structure, but at the cost of introducing extra parameters.

Mean-Shift [2], [4], [5], a very popular method[5], derives

a variant of the gradient ascent, written as

$$x(n+1) = \frac{\sum_{i=1}^{N} x_i K_\sigma(x_i, x(n))}{\sum_{i=1}^{N} K_\sigma(x_i, x(n))},$$

where $K_\sigma(x, y)$ denotes the kernel function (e.g., for Gaussian kernel: $K_\sigma(x, y) = \exp(-\sigma||x - y||)$ and $\sigma$ is the kernel bandwidth. Mean-Shift successfully solves the 1st problem, since it apparently does not involve the fixed step length parameter $\lambda$ and actually Mean-Shift is an adaptive steepest ascent in which the shifting step length can change adaptively with the local density distributions of dataset [4], [5]. Mean-Shift can guarantee to converge under the condition that the kernel is chosen appropriately [5] (i.e., the 4th problem is partly solved). Nevertheless, Mean-Shift apparently doesn't change the nature of numerical iteration of gradient ascent, and it is in effect equivalent to gradient ascent [4], [5]. For these reasons, Mean-Shift still suffers from the other problems of gradient ascent.

Besides Mean-Shift, there is also a graph-theoretical-based gradient ascent (GGA) [9]. GGA first defines the parent node for each node approximately[6] along the direction of gradient as follows,

$$I_i = \arg\max_{j \in \eta_\sigma^i} \frac{f_{j,i}}{d_{j,i}},$$

where $\eta_\sigma^i$ denotes the neighbors of point $i$ within its local radius $\sigma$, $d_{j,i}$ denotes the distance between points $j$ and $i$, and $f_{j,i} = f(x_j) - f(x_i)$ denotes the density difference between points $j$ and $i$. Then, GGA constructs a directed graph by linking each node to its parent node, and searches the root nodes (approximately the density peaks) on the constructed graph consisting of several unconnected subgraphs. Each sub-graph actually represents a cluster. Unlike Mean-Shift, the graph-theoretical implementation of GGA helps solve the 2nd and 3rd problems. Specifically, the time cost on searching the root nodes on the constructed directed graph could be negligible; each point in the dataset is viewed as a node in the graph, which is blind to the data attributes. However, despite avoiding the iteration step length $\lambda$, GGA requires a local parameter $\sigma$ to specify the neighbor nodes in the definition of the parent node and is sensitive to the parameter $\sigma$. Besides, GGA requires non-trivial effort to avoid cycles in the constructed graph (which lacks of strict proof in the paper); otherwise the search on the graph could not converge at certain root node.

In 2014, Rodriguez and Laio (RL) [10] proposes a simple yet powerful non-gradient density-peak-based clustering method. The novelty of their method is that an interactive way is designed to fast determine the density peaks (or cluster centers) based on an insightful observation (or discovery): the points in the density peaks are the ones with "a higher density than their neighbors and relatively large distance from points with higher densities" [10]. Specifically, they first re-define each point $i$ using two variables (or features) $\rho_i$ and $\delta_i$. $\rho_i$ denotes the local density, written as

$$\rho_i = \sum_{i=1}^{N} \varphi(d_{i,j} - \hbar),$$

---

1. The time-complexity is $O(N^2T)$, where $N$ denotes the size of the dataset and $T$ the number of numerical iterations.

2. This is due to the nature of the numerical iteration of the method.

3. The iteration may oscillate around the optimum, a well-known problem for gradient ascent.

4. An under-smoothed density surface could lead to the over-partitioning clustering result, while an over-smoothed density surface could lead to the under-partitioning clustering result.

5. Mean-Shift was first proposed in [2], developed in[4] and became popular since the work in [5]). Actually, the idea of Mean-Shift can be also found in literature [6], [7] where it is called the Iterative Local Centroid Estimation (ILCE), and DENCLUE-2.0 [8].

6. GGA approximates the gradient at each point $i$ by $\max_{j \in \eta_\sigma^i} \frac{f_{j,i}}{d_{j,i}}$.

where $\varphi$ is a self-defined function ($\varphi(x) = 1$ if $x < 0$ and $\varphi(x) = 1$ otherwise), and $\hbar$ is a cutoff distance parameter. $\delta_i$ is the smallest distance between point $i$ and the other points with higher densities $\delta_i$, written as

$$\delta_i = \min_{j:\rho_j > \rho_i} (d_{i,j}).$$

Particularly, for the point with the globally highest density, its distance variable is defined as $\delta_i = \max_{j \in X}(d_{i,j})$, where $X = \{1, \cdots, N\}$. Then, based on the above new representation, they plot all samples of the original dataset in a 2D Euclidean space, which results in a 2D scatterplot called the *Decision Graph*. In this Decision Graph, since the points in the density peaks are the only ones with relatively high values in both variables (as compared to the non-density-peaks points), they will pop out as outliers and can thus be interactively determined by users. Obviously, RL's method belongs to the General Hierarchical clustering (G-HC) (see details in Section 3.1), since RL's method seeks for a particular data representation (i.e., the Decision Graph) in the first phase, based on which an optimal flat partitioning of the original dataset can be obtained in the 2nd phase. RL's method has a very close relationship with the proposed method: the way of defining the variable $\delta_i$ corresponds to the method of Nearest Ascent; the Decision Graph can be viewed as a special data representation of determining the undesired edges in the In-Tree.[7] In other words, RL's method and the proposed method are in essence the same (i.e., a non-gradient density-peak-based clustering method), and RL's method is a special form of the proposed method. For this reason, like the proposed method, RL's method could also solve all the listed problems in the gradient ascent.

**Representative methods in the 2nd category: DBSCAN, OPTICS and HDBSCAN.** This category (the density-border-based) can be traced back to the work of Hartigan in 1975 [11], while the most popular and widely used algorithm is the one called DBSCAN proposed in 1996 [12]. DBSCAN[8] implicitly defines the density at each data point $i$ as the number of the $r$-neighboring[9] points of point $i$ within the radius centered at point $i$, and accordingly defines the points with densities beyond a pre-defined density threshold ($MinPts$) as the core points. For the non-core points (with densities below $MinPts$), they are further divided into two parts: the border points and noise points, which differs in that the border points are the neighboring points of certain core points. A cluster is then defined as a maximum set (denoted of $S1$) of core points in which any core point is the $r$-neighboring or transitively[10] $r$-neighboring point of any other core point in this set, plus a maximal set (denoted as $S2$) of the border objects which are the neighboring nodes of certain core objects in $S1$. DBSCAN has two major problems: (i) the two parameters, $r$ and $MinPts$, are hard to set; (ii) Using a global density threshold, DBSCAN could fail to detect the clusters with varying densities [13], [14], [15].

Among the various variants of DBSCAN (see a review in [15]), OPTICS [16] and HDBSCAN [17] are more representative.

Unlike DBSCAN which directly seeks for a flat partitioning of the dataset, clustering based on OPTICS (or OPTICS for short[11]) belongs to G-HC, which divides the clustering process into two distinguishable phases. In the 1st phase of G-HC, OPTICS first seeks for an effective linear arrangement of the samples in the dataset, via a walk on a minimal spanning tree based on the so-called reachability distances [15]. Then, similar to the Decision Graph [10], OPTICS represents each data point by two variables, its index in the new order and its smallest reachability distance, and then shows the original dataset in a 2D bar plot, called the **reachabiltiy plot**[12]. Actually, there is a close relationship between this reachability plot and the Dendrogram [18]. In the 2nd phase of G-HC, like Dendrogram, one can obtain a flat clustering result from the reachability plot by such as setting a global threshold (for the parameter $r$). Compared with DBSCAN, the advantages of OPTICS are threefold. (i) OPTICS is relatively insensitive to the two parameters, $r$ and $MinPts$, especially for $r$ which is usually set as infinity (in the first phase of G-HC). (ii) By choosing different distance thresholds[13] in the reachability plot, one can easily get all the flat clustering results obtained by[14] DBSCAN* with same values for $MinPts$ and $r$. (iii) OPTICS can discover the clusters in varying densities that DBSCAN cannot. The underlying reason is that in the 2nd phase one can choose other flexible methods [18] to detect the clusters in more complicated reachability plots in which simply setting a global threshold would fail. Besides, OPTICS also benefits another significant variant of DBSCAN, called HDBSCAN [17] (also a G-HC method).

HDBSCAN [17] was proposed recently in 2015, which follows the two phases utilized in OPTICS. However, compared with OPTICS which is overall not easy to understand, HDBSCAN presents a more concise and comprehensible procedure in the 1st phase to make the dataset organized into a condensed Dendrogram based on the concepts such as the minimum cluster size and cluster shrink.[15] In the 2nd phase, HDBSCAN proposes an effective automatic cutting method to extract the flat partitioning result from the con-densed Dendrogram. The cutting method is mainly based on the concept of cluster stability and the objective that that the sum of the stabilities of the extracted clusters should be the maximal. Theoretically, this automatic cutting method could extract the clusters in varying densities. Compared with OPTICS, HDBSCAN completely abandons the parameter $r$, and like OPTICS, HDBSCAN is not sensitive to the parameter $MinPts$ (the only parameter in HDBSCAN). However, this is at the cost of a time complexity of $O(N^2)$ (corresponding to the worst case of OPTICS when $r$ is set as infinity) [19].

---

7. Note that the points in the density peaks are actually the starts nodes of the undesired edge

8. Instead of introducing DBSCAN using the way as defined in the original paper which involves a serial of non-trivial definitions, here we will introduce it using a simple yet equivalent way.

9. Here we say that point $j$ is a $r$-neighboring point of $i$ if point $j$ is within the radius $r$ centered at point $i$ (note that this $r$-neighborhood relationship is symmetric. i.e., point $i$ is also a $r$-neighboring point of point $j$ with respect to $r$).

10. Here we say that point $j$ is a transitively $r$-neighboring point of $i$ if there is a chain of points $p_1, \cdots, p_n$ (where $p_1 = i$ and $p_n = j$) such that $p_k$ and $p_{k+1}$ (for any $k \in \{1, \cdots, n-1\}$) are the $r$-neighboring points to each other. Obviously, this transitively neighboring relationship is also symmetric.

11. Note that in the original literature OPTICS only refers to the 1st phase in which a hierarchical data representation is constructed. However, OPTICS here refers to the clustering based on OPTICS, which also contains the phase in which certain flat partitioning is extracted from the hierarchical data representation.

12. Each valley in the reachability plot represents a optimal cluster.

13. In conclusion, for the parameter $r$, it is usually set as infinity in the first phase so as to output a complete hierarchical data representation, and is set as a specific threshold in the 2nd phase so as to obtain an optimal flat partitioning of the original dataset.

14. DBSCAN and DBSCAN* differ in that the latter one only takes the core points as the elements of a cluster

15. A good tutorial is given in http://hdbscan.readthedocs.io/en/latest/how_hdbscan_works.html.

**Comparison between the two categories.** Compared with the 1st category (i.e., density-peak-based) of density clustering, the 2nd category (i.e., density-border-based) is more capable of detecting arbitrary shapes of clusters, despite the fact that (i) some methods (e.g., RL's method) in the 1st category can detect certain non-spherical clusters that K-means cannot and (ii) some methods (e.g., DEN-CLUE) claim to be able to detect arbitrary shapes of clusters.[16] The reason is that the 1st category has an implicit assumption for cluster centers (actually being the density peaks) that the 2nd category doesn't have, and thus the 1st category could have the risk of over-partitioning the clusters that are not featured with the density peaks, instead, the ridges [20], [21]. Compared with the 1st category, the 2nd category uses a relatively poor neighborhood-based density approximation[17]. Take DBSCAN for instance. It is usually viewed as a special case of DENCLUE using the simple uniform spherical kernel [15], [3]. For this reason, the density-peak-based methods could perform better than the density-border-based methods in certain applications, since they could benefit from all kinds of density estimation methods.

## S2 ABOUT THE DATA INDEX

In the original formula of Nearest Descent (see the 4-th step of stage I), the data indexes (i.e., the order in which data points are processed), together with the potential values, serve to provide references for parent node selection. The role of the data indexes is to deal with the special case that the close points (or the extreme case, the overlapping ones) could have the same potential. Specifically, the data index term could help the proposed method to choose only one point from those close points to communicate with other points with lower potentials, which can avoid the unexpected "*singleton cluster phenomenon*" after edge cutting.

However, introducing data index term does not mean that the clustering result is sensitive to the change of data indexes. The influence is very limited, due to the following reasons:

16. Note that DENCLUE actually requires post-processing steps to merge the over-partitioning clusters, which is at the cost of introducing an extra parameter.

17. Specifically, DBSCAN is based on $r$-Nearest-Neighbor criterion, whereas OPTICS is based on $k$-Nearest-Neighbor criterion, where $k$ refers to the parameter $MinPts$. Actually, this difference implicitly happens at the time when OPTICS introduces the **core distance** (later inherited by HDBSCAN). For this reason, one can also notice that the roles of the two parameters, $r$ and $MinPts$, in DBSCAN and OPTICS actually differ.

If point $i$ is not with the local minimal potential, then since its candidate point set $J_i$ consists of two groups of points, i.e., the points with smaller potentials (group A) and the points with the same potential but smaller data indexes (group B), only points in group B are affected by data index. The points in group B are so special that they do not always exist. Even if group B is non-empty, according to the "constrained proximity criterion" claiming that only the nearest point or the point with the smallest index among the nearest ones in $J_i$ will be selected as the parent node, the selected parent node $I_i$ is at least very close to point $i$. Therefore, the change of the data indexes only affects the connections or edges in local areas, whereas these local edges, even they are changed, are in general much shorter than those undesired edges among groups and thus less likely to be cut off. Consequently, the near points connected by these local edges can be clustered into the same groups.

Otherwise, the candidate points with lower potential values in set $J_i$ will come from other groups. So, no matter how data indexes are assigned to those points, the edge between point $i$ and its parent node $I_i$ will in general be much longer than the edges among points within group, and thus will be cut off later. This is what we expect. Consequently, the clustering result will not be largely affected. The above analysis is further demonstrated by a test with different random permutations to dataset "S1", as shown in Fig. s8.

## S3 PROOF OF THE IN-TREE

Before proving, some definitions and properties [22], [23] about the graph are listed as follows.

- The indegree $d^-(v)$ and outdegree $d^+(v)$ of a node $v$ refer to the number of directed edges ended at it and started from it, respectively. The degree $(d(v))$ of a node $v$ refers to all the directed edges connected with it, including the directed edges both ended at and started from it.
- A self-loop at a node is a directed edge that points to the node itself.
- A directed cycle is such that for any node on it, traveling along the edge directions, it can return to itself.
- A connected graph is a graph of which any two nodes are connected by a sequence of edges.
- A digraph is a graph with the edge being directed.

The graph $\hat{G} = (\hat{V}, \hat{E})$ has the following properties:

(P1) If $\hat{G}$ is a cycle, then, for any node $v \in \hat{V}$, $d(v) = 2$. Typically, when the cycle is a directed cycle, $d^-(v) = d^+(v) = 1$.

(P2) If $\hat{G}$ is a tree, then $m(\hat{G}) = n(\hat{G}) - 1$, where $m(\hat{G})$ and $n(\hat{G})$ denote the number of edges and the number of nodes, respectively.

(P3) If $\hat{G}$ is a digraph, then $d(v) = d^-(v) + d^+(v)$, and $\sum_{v \in V(\hat{G})} d^+(v) = \sum_{v \in V(\hat{G})} d^-(v) = m(\hat{G})$.

**Proof.** First, the definition about the graph $G = (V, E)$ can be concluded as the following requirements:

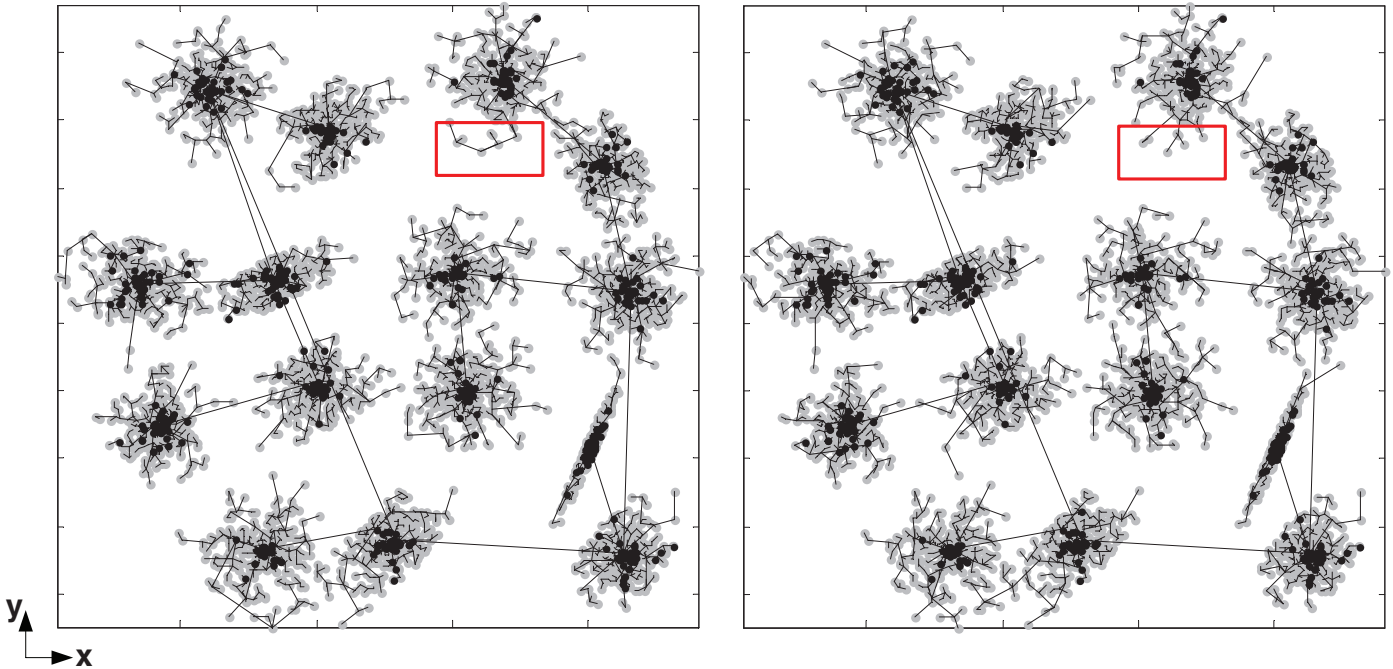

Fig. s8. The experiment for showing the negligible influence of data index. Take dataset "S1" for instance. In order to show the effect of data index, we first generated lots of vertices with identical potentials, since only for these vertices, their data index may serve as a reference for identifying parent nodes. We set $\sigma$ as 10, much less than the average pair-wise distance. Consequently, only 151 potential values are unique, which means that most vertices have the same potentials. We randomly changed the data index for 50 times, and the In-Trees of two of the 50 runs are shown in this figure. It is clear that for the edge points in the red rectangles, local difference occurs in their neighborhood relationships, whereas this difference makes very little effect on the final clustering results. In fact, (i) for the in-trees of all 50 tests, we found that the longest 100 edges are the same; (ii) for the clustering results of all 50 tests, when we took one (e.g., the first clustering result) as the benchmark and compared it with the other 49 clustering results, relative clustering error rate were very small, $0.0074 \pm 0.0041$ (mean $\pm$ standard deviation), which means that on average, 37 points out of 5000 get different assignments. To summarize, although the nearest descent rule involves data index item, the change of the data indexes makes negligible influence on the final clustering result.

(R1) For each node $i$ in $G$, its $J_i$ are defined as the points with smaller potentials or the same potential but smaller data indexes;

(R2) For each node $i$ in $G$ with $J_i \neq \varnothing$, its (true) parent node $I_i$ is unique, i.e., $d^+(i) = 1$;

(R3) For each node $i$ in $G$ with $J_i = \varnothing$, it has no parent node, i.e., $d^+(i) = 0$;

The following proof is then made by showing how the requirements (R1~R3) can meet the definitions (a~d) of the In-Tree in Section 2.4 of the main body of this paper.

(i) Let $\hat{P}$ be the globally minimal potential among the potentials at all nodes, i.e., $\hat{P} = \min_{i \in X} P_i$. Obviously, according to R1, for each node $i$ with $P_i < \hat{P}$ where its $J_i \neq \varnothing$. If only one node has the minimum potential value, then, according to R1, $J_i = \varnothing$. If several nodes are of the minimum potential, only one node among them having the smallest index (note that the node index is unique) has empty $J_i$. In conclusion, there is always one and only one node, denoted as $r$, with $J_r = \varnothing$, for which, according to R3, $d^+(r) = 0$. **The condition (a) is met.**

(ii) According to R1, for any other node $v(\neq r)$, there will be at least one node (e.g., node $r$) which makes $J_v \neq \varnothing$. Then, according to R2, we have $d^+(v) = 1$. **The condition (b) is met.**

(iii) Suppose there is a cycle $C$ in digraph $G$. If $C$ is a directed cycle $v_i \rightarrow v_j \rightarrow \cdots \rightarrow v_i$, then according to R1, the potential values for these nodes should meet $P_i \geq P_j \geq \cdots \geq P_i$, which is true only when $P_i = P_j = \cdots = P_i$. Then according to R1 again, the data indexes for these nodes should meet $i > j > \cdots > i$, resulting in $i > i$, an obvious contradiction. If $C$ is not a directed cycle, we claim that there exists at least one node of $C$, say $w$, then, according to P1, there will be either $d^+(w) = 2$ or $d^-(w) = 2$ in this cycle $C$. If $d^+(w) = 2$, then it will contradict with the result in (ii); If $d^-(w) = 2$, since every node $v$ on a cycle has degree 2, i.e., $d^+(v) + d^-(v) = 2$, and according to P3,

$$\sum_{v \in V(C)} d^+(v) = \sum_{v \in V(C)} d^-(v), \tag{1}$$

there must be at least one node in $C$ having outdegree 2, which contradicts the result in (ii) again. Therefore, there is no cycle in digraph $G$. **The condition (c) is met.**

(iv) Suppose $G$ is not connected, e.g., containing $M$ connected sub-graphs $G_1$, $G_2$, $\cdots$, $G_M$. Since digraph $G$ has no cycle, there should be no cycle in each subgraph as well. Hence, each subgraph $G_k$ is at least a tree. Assume the node $r$ of outdegree 0 is in $G_i$, then nodes in any other subgraph $G_k$ ($k \neq i, k \in \{1, 2, \cdots, M\}$) are all with outdegree 1, thus, according to P3,

$$m(G_k) = \sum_{v \in V(G_k)} d^+(v) = \sum_{v \in V(G_k)} 1 = n(G_k), \quad (2)$$

which contradicts P2. Therefore, the digraph $G$ is connected. **The condition (d) is met.**

According to (i), (ii), (iii) and (iv), we can conclude that *the digraph $G$ is an In-Tree*.

## S4  MATLAB CODE FOR ND BASED ON M-CUT

In the following, we give a self-defined function that could implement the proposed method. For the input variables of the function, "D" denotes the distance matrix, "M" the undesired edge number, and "sigma" the parameter (i.e., the kernel bandwidth) in potential computation. For the output of the function, "Label" denotes the cluster assignments of all the samples. Note that, 1) the vector "I" stores the parent nodes of all nodes (e.g., "I(i)" stores the parent node of the node $i$); the vector "W" stores the edges lengthes (e.g., "W(i)" stores the edge length from node $i$ to node "I(i)"); the vector "P" stores the potentials at all nodes (e.g., "P(i)" stores the potential at node $i$). 2) We set the length of the self-loop added at each root node as the negative infinity ("-inf"). 3) To fully avoid the impact of index, one could comment out the 4th line.

```
function Label=ND(D, M, sigma)
% Stage 1: construct the IT
S=exp(-D./sigma); P=-sum(S,2); N=length(P);
C=repmat(P',N,1)-repmat(P,1,N);
D(C>0)=inf;
D((C==0)&(tril(ones(N,N))))=inf;
[W,I]=min(D,[],2);
idx=find(isinf(W)); I(idx)=idx; W(idx)=-inf;
% Stage 2: remove the undesired edges
for i=1:M, [u,idx]=max(W); I(idx)=idx; W(idx)=-inf; end
% Stage 3: gather at the root nodes
I_old=(1:N)';
while norm(I-I_old), I_old=I; I=I_old(I_old); end
% Stage 4: cluster assignment
idx=find(isinf(W)); Label=zeros(N,1);
for i=1:length(idx), Label(I==idx(i))=i; end
```

## S5  CUTTING DIRECTLY BASED ON THE IN-TREE

As shown in Fig. s6, the data representations EP, SP, DE and BP are derived from the initial data representation, i.e., the In-Tree. Actually, besides on the directed data representations, one can remove the undesired edges directly based on the In-Tree, which, however, has additional requirements for the dimension of the dataset and the labels. Specifically, one can interactively remove the undesired edges on the In-Tree if the dimension of the test dataset is no larger than 3, since in this case the constructed In-Tree can be visualized in a 2D Euclidean space. Due to the distinguishable features of the undesired edges, it is very easy to design reliable method for removing them interactively. See details in Fig. s9. This interactive method could also be of a broad meaning, since there are many effective dimensionality reduction methods, such as principle component analysis (PCA) [24], multi-dimensional scaling (MDS) [25], locally linear embedding (LLE) [26], ISOMAP [27], t-Distributed Stochastic Neighbor Embedding (t-SNE) [28] and Tree Preserving Embedding (TPE) [29], that can be used to embed the high-dimensional dataset into the 2D space. Nevertheless, the embedded datasets may suffer from the so-called "crowding problem" [28], [29], that is, some clusters may become overlapped in the embedding, which would to some degree reduce the reliability of the clustering result. For instance, Fig. s9d shows the In-Tree of the embedded "Mushroom" dataset. Although the clustering assignment in Fig. s9e is quite consistent with our visual perception, it actually achieved a purity of 0.997 rather than 1, due to the overlapping of the clusters in the embedding.

Besides, if users can label a small number of samples in the datasets, the above constraint for the dimension of the dataset could be eliminated. The reason is that, due to the characteristics of the In-Tree and the undesired edges, it is very easy to perform the semi-supervised cutting according to the cell-division-like cutting criteria[18]: (i) the impure sub-trees should be divided into pure ones, where "impure sub-tree" refers to the one containing different kind of labeled nodes, and "pure sub-tree" refers to the one containing the same kind of labeled nodes; (ii) for each impure sub-tree, the edges will be explored in decreasing order, whereas whether the explored edge should be removed depends on whether each generated sub-tree after removing certain edge contains at least one labeled node. Thus, the goal of this cell-division-like supervised cutting is to make all the generated trees become pure, as illustrated in Fig. s10a. Note that in the end, the pure sub-trees with the same labeled data will be merged as one larger pure sub-tree. This semi-supervised cutting is free from the constraint for the dimension of the dataset; the whole process is automatic; the cluster number is guaranteed to be equivalent to the class number of the labeled data. Besides, this semi-supervised cutting could perform the local cutting. Specifically, when handling with the datasets "S1&0.01S1" which contains different scales of clusters, the undesired edges between the large-scale clusters and the ones between the small-scale clusters can all be determined and removed, as proven in Fig. s10b. In general, more labeled points could lead to more reliable results.

The effectiveness of this supervised cutting is also demonstrated on the datasets "Mushroom" and "Olivetti Face"[19]. For the dataset "Mushroom", it should previously be rather tedious for scientists to label 8124 mushrooms as either poisonous or edible one by one, whereas Fig. s10c reveals that scientists only need to label a few number of mushrooms. For the remaining ones, satisfactory prediction can be made by the proposed method. And the more the mushrooms are labeled, the more reliable the prediction could be. The Dataset "Olivetti Face" is regarded as a hard dataset due to its extreme sparsity [10], whereas the result in Fig. s10d shows that considerable performance could be achieved if computer could "view" in advance the faces of each subject several times. Note that although the labeled data can also benefit K-means for its initialization (i.e., K-means can take those labeled points as the starting points or seeds), the inherent drawbacks (e.g., the drawback in detecting the non-spherical clusters) of K-means make this benefit not always as obvious as the proposed method.

---

18. One can see details in arxiv: 1412.7625.

19. Olivetti Face dataset contains 400 grayscale face images (112 × 92 pixels) from 40 subjects. Each face can be represented by vector of 10304 features.